# Where in the Input Does a One-Pass Decision Encoder Look?

*Evidence position, distractors and the 512-token cut in Laya, with a window-size guide*

Mohammed Nayeem Khan
mohammed.nayeem.khan@intel.com
mnknayeem@gmail.com

Allan Kevin T
allan.kevin.t@intel.com

Intel Corporation

**Abstract**

Laya is a small model that answers a multiple-choice question about a text in one pass: the question, the options (each with a name and a description) and the text form one sequence of at most 512 tokens, the window, in the English release. A longer text is silently cut at the end, and we measure what that costs. Test documents place a held-out row's text, the evidence, at a chosen token offset among distractor text; 177 items, five tasks, three checkpoints (released, short-row fine-tune, long-document fine-tune), four window sizes and 93,275 scored requests. When the evidence lies beyond the window, accuracy falls to 26.4% for the released checkpoint (27.2% for both fine-tunes), near the 27.0% of guessing and below the 35.6% of always giving the commonest label. Confidence gives no warning (AUROC 0.50 to 0.55 for telling cut from read requests; 0.5 = chance), and the runtime's truncation flag also fires on requests whose evidence was visible. The released checkpoint reads visible evidence badly once text precedes it: 38.5% pooled against 60% when the evidence comes first, so a larger window alone gains +6.2 points. Fine-tuning on long documents adds +27.6 points on visible evidence over the short-row fine-tune (95% interval +22.3 to +33.1) and nothing on cut evidence. That is an in-family result: training rows and test documents share generator and question types, and the main filler is of the training kind. With encyclopedic filler never seen in training the gain is +13.2 (+9.9 to +16.6), with rows of other tasks as filler -2.0. Among training-free rearrangements a 312-token window scan helps most (+11.9 points over the default for the released checkpoint) at six to eleven forward passes. Documents are synthetic, labels are LLM-made and one backbone is studied, so these diagnose this model, not long documents in general.

### Key findings

1. The cut is silent: cut-evidence accuracy 26.4% / 27.2% / 27.2% (released / short-row / long-row; a uniform guess gives 27.0%), mean confidence 0.57 when cut against 0.57 when read (released).
2. Confidence cannot find it: AUROC 0.50 / 0.53 / 0.55; at a 0.9 gate the long-row checkpoint accepts 59% of cut requests with 65% wrong.
3. Inside the window the released checkpoint is distracted, not blind: 60% with the evidence first, 38% after 100 tokens of boilerplate, even at `max_len` 2048.
4. Window and training work together: raising `max_len` from 512 to 2048 gains +6.2 / +3.4 / +21.6 points (released / short-row / long-row). Long-document training adds +27.6 points on visible evidence and +0.1 on cut evidence, and nothing when rows of other tasks are the filler. The short-row fine-tune is within noise of the released model (+1.9 points, interval -4.4 to +9.1).
5. With two competing pieces of evidence the later one wins when they are adjacent, and the pieces are about even at gaps of 100 to 400 tokens (sentiment, released). A window scan beats a 2,048 window by +7.8 points (released) and +7.0 (short-row), the two checkpoints that cannot use a long window.
6. Costs: a larger `max_len` is free for short inputs; for a window-filling batch of eight the latency grows from 206 ms at 512 to 2971 ms at 4,096 on a shared GPU.

# 1 Introduction

A one-pass decision encoder answers a multiple-choice, scoring or yes/no question about a piece of text with a single forward pass: question, options and text go in as one token sequence and probabilities over the options come out, with nothing generated. Laya is a public model of this kind, an English ModernBERT-large encoder with a two-layer option head. It is cheap and fast, and its probabilities are meant to be thresholded, but it reads a fixed-length input. The released configuration reads at most 512 tokens in total, and when the text is longer the *end* of the text is not read. This paper asks where in the input the model looks and how sharply it stops looking. It also asks whether anything in its output reveals that evidence was dropped, and what the cheapest way to avoid the loss is.

The cut is silent. The runtime records a `truncated` flag and the number of state tokens it kept, but the probabilities that come back are those of a model that has seen a shorter document: not flatter and hardly lower, and nothing marks them. In our synthetic corpus 0.0% of the 1,221 held-out rows are longer than the 485 tokens that fit next to the lightest question head at the default setting, and 0.0% are longer than what fits next to the heaviest head. Real traffic will differ, so a user needs the same measurement on their own data (a recipe is in Section 10).

We take held-out rows from five decision tasks (sentiment, hardware failure, urgency, a 12-way topic and a department / churn pair), keep each row's text as the *evidence*, and surround it with programmatically generated distractor text so that the evidence sits at a chosen position of a longer document. The label is the row's own, unchanged by construction. The distractor text carries no label signal for the question asked. Three checkpoints (the released model and two fine-tuned ones) are scored at four `max_len` settings (512, 768, 1024, 2048). We vary where the evidence sits and how much distractor text there is, and of what kind. We also test whether two pieces of evidence compete and which no-training truncation strategy replaces the default, and we measure confidence on cut documents and the cost of a larger `max_len`.

Two separate losses show up. First, with the default setting, a request whose evidence lies entirely beyond the window is answered correctly 26.4% [24.4-28.5] of the time (released checkpoint, n=1,770). That is close to the 27.0% of a uniform guess among the options and below the 35.6% of always answering each question's commonest label; the two fine-tuned checkpoints score 27.2% and 27.2% (Section 5, Table 8). Second, even where the evidence is inside the window, the released checkpoint reaches only 38.5% [37.2-39.8] (n=5,307). With the evidence at the very start of the document it scores 60.5%; with 100 tokens of boilerplate in front of it, 38.4%, and a window of 2,048 tokens does not rescue it. A checkpoint fine-tuned on long documents reads the same visible evidence at 68% [67-70] and shows a sharp cliff at the end of the window, which the other two do not because they have little accuracy left to lose. Section 4 places the cut, Section 6 shows that confidence does not reveal it (AUROC 0.50), and Section 7 separates what retraining adds from what a configuration change gives for free.

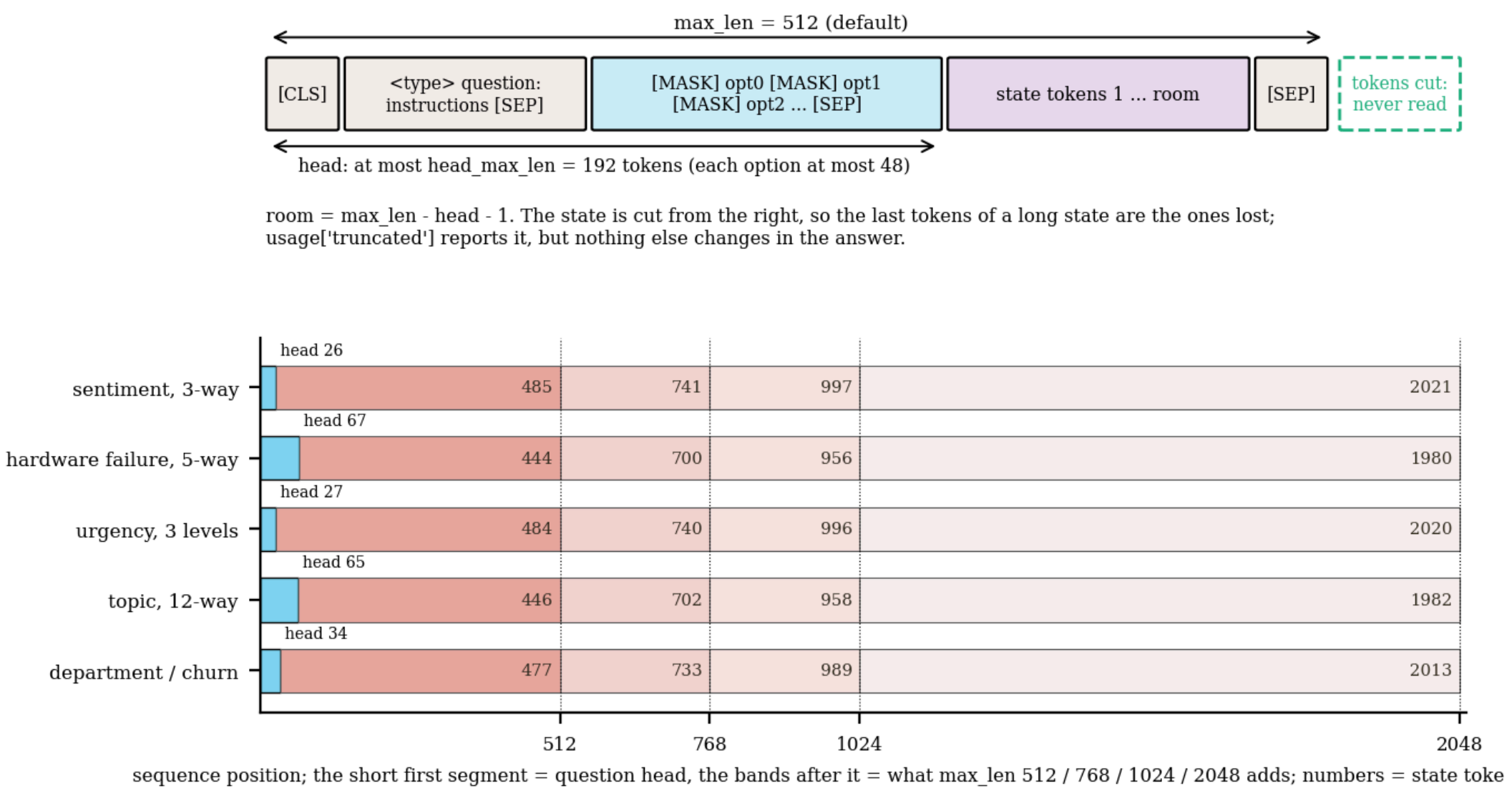

**Figure 1.** What the encoder reads. (a) The token sequence of one call: a head made of the question and every option behind its own [MASK] slot, the state, then separators. (b) To scale, for each of the five questions used below: the head (the short first segment of each bar) and how many state tokens fit when `max_len` is 512, 768, 1024 or 2048 (numbers).

**Table 1.** State tokens that fit (the 'room') for each question and `max_len`; head = tokens spent on instruction and options. The department / churn task pools two question schemas; the row shows the 2-way one (the 3-way one has a head of 37 tokens).

| Question | Options | Head | max_len 512 | max_len 768 | max_len 1024 | max_len 2048 |
|---|---|---|---|---|---|---|
| sentiment, 3-way | 3 | 26 | 485 | 741 | 997 | 2021 |
| hardware failure, 5-way | 5 | 67 | 444 | 700 | 956 | 1980 |
| urgency, 3 levels | 3 | 27 | 484 | 740 | 996 | 2020 |
| topic, 12-way | 12 | 65 | 446 | 702 | 958 | 1982 |
| department / churn | 2 | 34 | 477 | 733 | 989 | 2013 |

### 1.1 What is and is not new

That a fixed input window loses what lies outside it is textbook knowledge, and positional effects in long-context language models are documented in detail (Section 12). The new part is a measurement on one widely downloadable decision encoder. It covers where the window ends for each question and how sharp the transition is, and how it interacts with distractor type, length and competing evidence. It also tests whether the model's own confidence or truncation flag gives a usable warning and whether fine-tuning on long inputs gives anything that raising `max_len` alone does not. Finally, it measures what raising `max_len` costs in time and memory and compares three training-free workarounds. We found no earlier study of this kind for typed-decision encoders (Section 12 lists what we checked and its limits).

> **Scope.** All text is machine-made: the labels come from an LLM (Qwen3.8-27B, with a blind second pass; a different LLM, gemma4-31b, agrees with the stored label on 94-99% of rows depending on the set) and the long documents are programmatic concatenations, not naturally long texts. Conclusions about *where the window ends* are mechanical and should transfer; conclusions about *how much a model is distracted* by real long documents should not be read off these numbers. Per-cell sample sizes are modest (about 36 items per task); intervals are shown throughout.

## 2 How the model reads, and where it stops reading

The model packs everything it needs into one sequence. A short header names the question type and gives the instructions. Every option then follows behind its own mask token so the encoder can attend from each option slot to the whole sequence. The text to be judged (the *state*) comes last. A two-layer head scores the option slots, and a small calibration table turns the scores into probabilities. Nothing in this path looks at tokens that are not in the sequence, which is why the length budget matters.

```
[CLS] <type> question: <instructions> [SEP]
[MASK] <option 0 text> [MASK] <option 1 text> ... [MASK] <option k-1 text> [SEP]
<state tokens, at most room of them> [SEP]
```

The runtime first measures the header. An option longer than 48 tokens is shortened to 48, and the header as a whole is limited to `head_max_len` (192 in the released configuration). What is left of `max_len` after the header and one closing separator is the *room* for the state, and the state is cut from the right when it does not fit:

```
head   = tokens of the first two lines          # each option text capped at 48 tokens, whole head capped at head_max_len =
192
room   = max_len - len(head) - 1               # one slot is kept for the closing separator
state  = state_ids[:room]                      # right-hand cut: everything after the first `room` tokens is dropped
if room < 1: raise ValueError                   # the runtime refuses a max_len that cannot hold the head
```

The five questions in this study have heads between 26 and 67 tokens, so at the default `max_len` of 512 the room is 444-485 tokens (Table 1). A question with many long option descriptions can have a head close to the 192-token cap, which leaves as little as 319 tokens for the text. The cut is by tokens, not words: one English word is roughly 1.3 tokens in this corpus, so a room of 485 tokens holds on the order of 350-400 words. Code, tables or non-English text consume the room faster.

Each answer comes with a usage record that includes the number of state tokens kept and a `truncated` flag. Probabilities are computed from the shortened sequence and nothing else: the flag changes nothing in the output, and no warning is raised. A `max_len` too small to hold the header raises an error, so the failure that occurs is "a little less room than the text needs".

The encoder underneath, ModernBERT, was pre-trained with a native sequence length of 8,192 tokens (Warner et al., Section 12), so 512 looks like a choice of the released configuration and not a limit of the encoder. In Laya the value is written into the checkpoint's configuration file and can be overridden for a single call, which is how all experiments below vary it without reloading anything. Whether the *trained head* behaves well at lengths it never saw in fine-tuning is an empirical question; Sections 4 and 7 test it.

### 2.1 Predictions E1 to E7

The layout above implies the following predictions, which the experiments check one by one (the verdicts are collected in Table 18).

- **E1.** Accuracy is a step function of the evidence position: close to the in-window level while the evidence is inside the room, close to the evidence-free level beyond it, with a transition whose width is about the evidence length.
- **E2.** The step moves with `max_len` by the same number of tokens, and differs across questions by the difference in head size.
- **E3.** Confidence does not drop after the cut, because nothing in the sequence is aware that anything is missing.
- **E4.** Distractor text *inside* the window costs little for neutral boilerplate and more for text that resembles the task; where two pieces of evidence disagree the first one wins whenever the second is cut.
- **E5.** Raising `max_len` costs time roughly in proportion to the tokens actually read, and almost nothing when the input is short.
- **E6.** Fine-tuning on long inputs helps only if the baseline cannot already use the longer window.
- **E7.** Re-arranging the text before the model sees it (keeping the end, keeping head and tail, selecting sentences, or scanning windows) recovers some of the loss, but is no better than a larger window.

## 3 Models, test material and protocol

### 3.1 Three checkpoints

Laya as released (called *shipped* below) is the public English checkpoint with `max_len` 512. The other two are fine-tunings of it made with Laya's own training command; only the training rows and, for the long-row one, the training `max_len` differ (the library also lists a typed-decisions checkpoint configured for 1,024 tokens, which we did not use). All three share one tokenizer, one head size and one configuration format.

- The **v1** checkpoint was trained on 2,818 generated rows with short states (median 34 tokens), for four epochs, at `max_len` 512.
- The **v1_long** checkpoint was trained on the same rows plus 761 *long* copies of a seeded 30% of the choice rows, in which the original text is embedded in 100-900 words of unrelated office boilerplate at a random position. It was trained at `max_len` 1024 (3,579 rows, longest 1189 state tokens; 13.4% of its rows exceed 485 tokens against 0.0% for v1).

A large language model (Qwen3.8-27B) wrote short cases conditioned on a target label for a fixed set of question schemas. A second pass of the same model, blind to the target, labelled each case again. Cases where the two disagreed were dropped. The held-out sets were generated the same way but separately, and de-duplicated against the training rows. An independent model (gemma4-31b) re-labelled the held-out sets and agrees with the stored label on probe 97%, sent 99%, json 97%, hw 95%, many 94%, ml 97%, urg 95%. Agreement of two LLMs is a filter and not ground truth, so the accuracies below measure agreement with the labelling model.

All fine-tuned checkpoints use the trainer's default combined probabilistic loss ('rlcd' in its configuration, with weights 0.75 and 1.0 on its two terms). The encoder learning rate is 2.5e-5 and the head learning rate 1e-4, with an effective batch of 64 (micro-batch 8 and 8 accumulation steps for v1, 4 and 16 for v1_long), bf16 and shuffled option order for choice questions. Both ran four epochs from seed 0, so v1_long saw about 27% more rows, namely copies of rows that v1 also saw. The trainer holds out 10% of the rows (at most 400) for calibration and abstention thresholds. Each checkpoint was trained on one consumer GPU (an RTX 4090).

### 3.2 Test material is built, not collected

We need documents in which the position of the decisive text is known exactly, so we construct them. The *evidence* is the state of one held-out row. Its label is the row's own label, because we never alter the evidence and the added text carries no label for the question asked. The *distractor stream* is assembled from hand-written sentence pools: 62 sentences of office and notice boilerplate (room bookings, badge rules, expense templates), 30 encyclopedic statements, the states of rows from *other* tasks, and, for sentiment, neutral-sentiment rows of the same task. No sentence of the pools is used in training data for any checkpoint; they are disjoint from the filler used to build the long training rows. A seeded shuffle gives each item its own stream, so two items never share a document except by design.

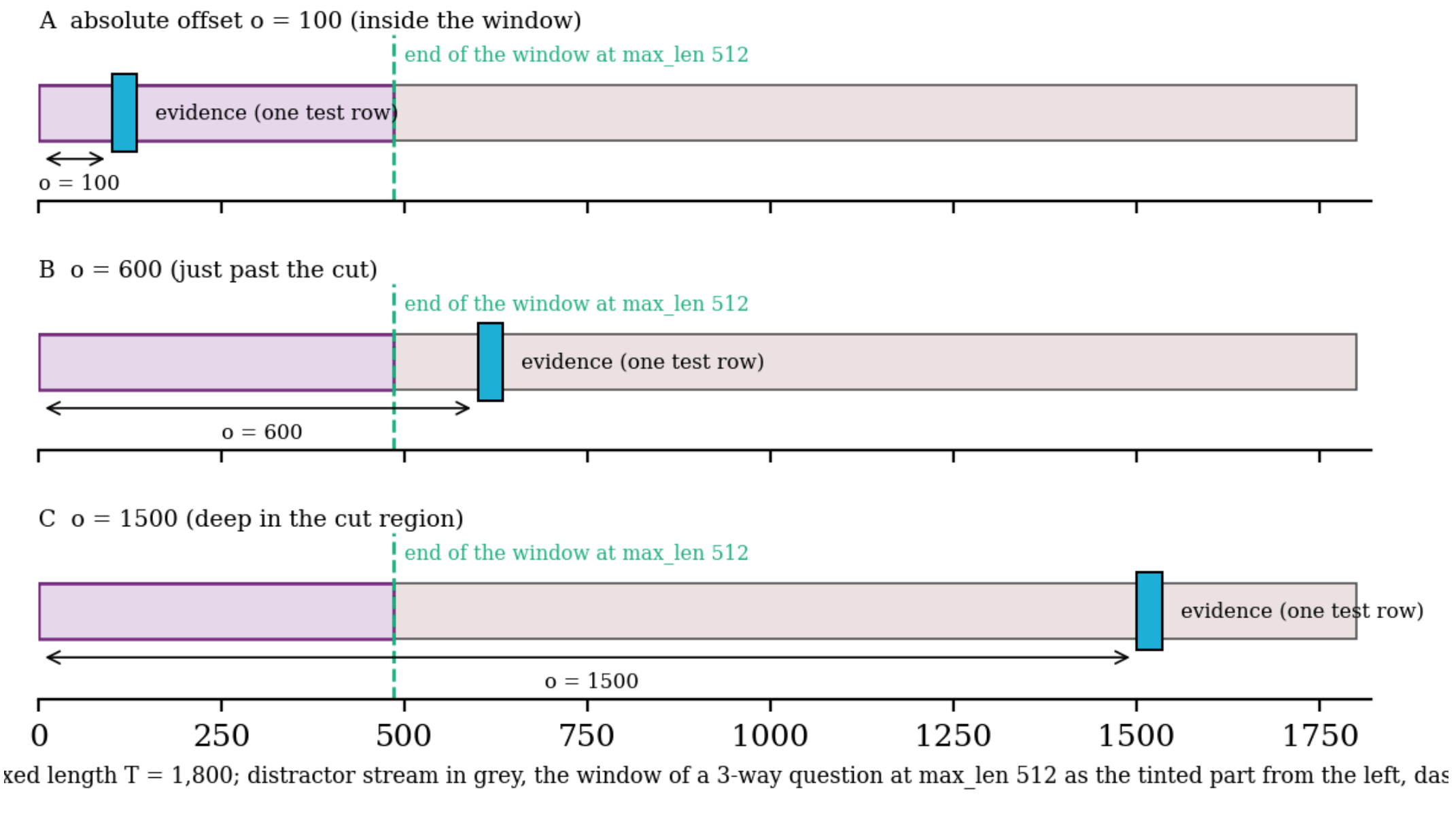

**Figure 2.** How one test document is made: the evidence (one held-out row, the small filled block) is placed at token offset o inside a stream of distractor text of fixed total length; the tinted part from the left is what a 3-way question reads at `max_len` 512.

Position is reported in two ways, because the two answer different questions. The first is the *absolute token offset* o of the first evidence token from the start of the state (0, 100, ..., 1500) at a fixed document length, which locates the cliff in tokens. The second is the *fraction* of the document at several document lengths, which shows whether the model prefers the beginning or the end. Tokens are counted with the model's own tokenizer, with character offsets mapped to token indices, so "evidence cut" means that none of its tokens survive in the window. "Partly visible" and "fully visible" are defined the same way.

The study uses five tasks: 36 sentiment, 36 hardware, 36 urgency, 36 topic (12-way) and 33 department/churn items, evenly spaced out of the held-out sets (177 items). The full sets hold 78-106 rows, and we subsample because the GPU was shared with other jobs. Requests whose visible prefix is identical (for instance, evidence-free tails that are cut anyway) are scored once and shared, which does not change any number.

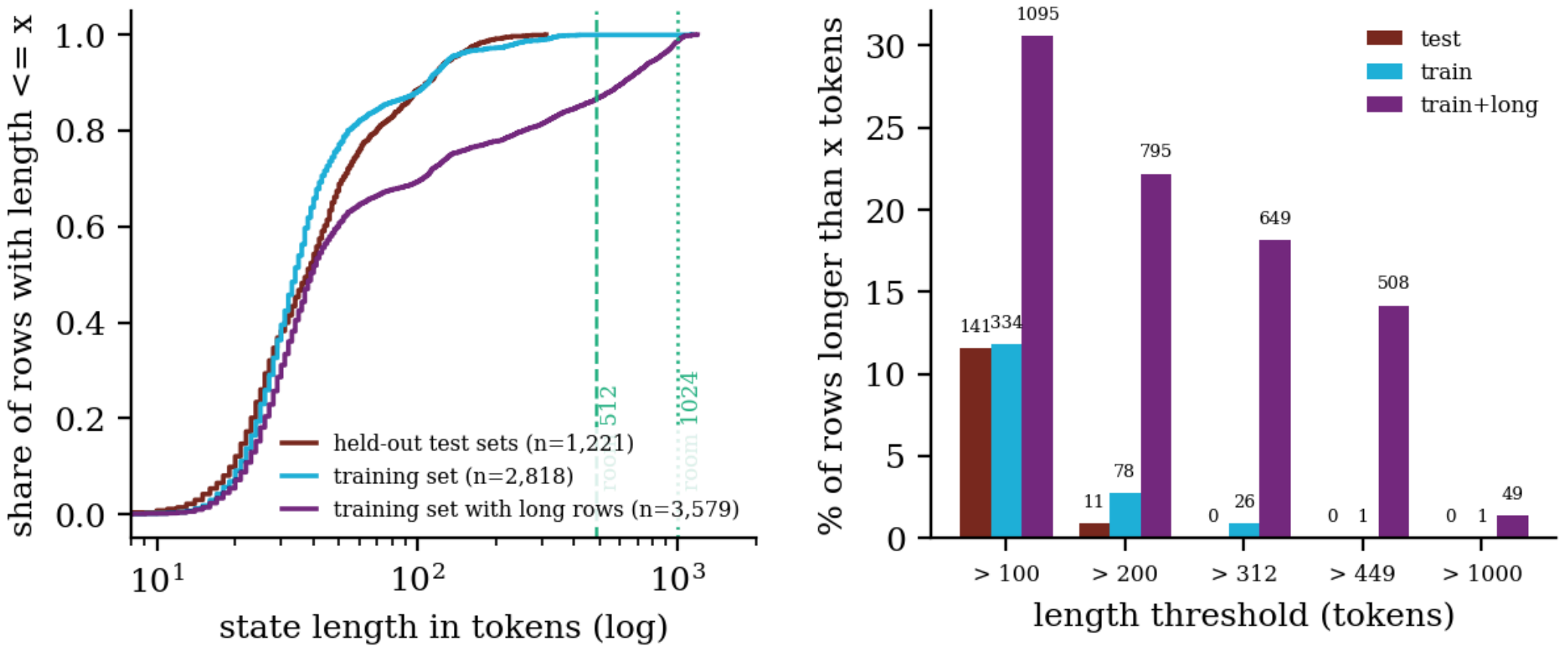

**Figure 3.** State length of the rows in the corpora. Left: cumulative share of rows up to a length, against the room at `max_len` 512 and 1024. Right: share of rows longer than a given number of tokens (counts printed on bars).

**Table 2.** State length of every data file in the corpus (tokenizer of the encoder). 'Beyond' columns give the share of rows whose state is longer than the room at `max_len` 512 under the heaviest head (319 tokens) and the lightest head (485), and at 1024 (831 and 997).

| File | Rows | Median | P95 | Max | >319 | >485 | >831 | >997 |
|---|---|---|---|---|---|---|---|---|
| sent_test | 106 | 20 | 28 | 33 | 0.0% | 0.0% | 0.0% | 0.0% |
| hw_test | 82 | 24 | 35 | 41 | 0.0% | 0.0% | 0.0% | 0.0% |
| urg_test | 78 | 34 | 55 | 71 | 0.0% | 0.0% | 0.0% | 0.0% |
| many_test | 195 | 34 | 59 | 73 | 0.0% | 0.0% | 0.0% | 0.0% |
| mix_test | 156 | 36 | 68 | 82 | 0.0% | 0.0% | 0.0% | 0.0% |
| json_test | 58 | 118 | 159 | 180 | 0.0% | 0.0% | 0.0% | 0.0% |
| ml_test | 322 | 62 | 168 | 211 | 0.0% | 0.0% | 0.0% | 0.0% |
| probe | 224 | 33 | 141 | 312 | 0.0% | 0.0% | 0.0% | 0.0% |
| train | 2,818 | 34 | 134 | 1131 | 0.9% | 0.0% | 0.0% | 0.0% |
| train_long | 3,579 | 39 | 831 | 1189 | 17.8% | 13.4% | 4.9% | 1.4% |

The corpus itself is mostly short: the median state has 38 tokens, the 95th percentile 136, the longest 312. Exposure to the default cut is therefore small (0.0% of test rows beyond the lightest-head room). Ordinary testing on short synthetic rows would never reveal the failure, which is why we build long documents deliberately.

## 3.3 What training and test share

There is no out-of-family test in this paper. The evidence in every test document is a held-out row of the same generator (Qwen3.8-27B) and the same question schemas that produced the training rows. The option names and descriptions are the same, and so is the labelling model. Held-out rows are de-duplicated against the training rows. The long-document fine-tune v1_long is trained on rows from that generator and those schemas, embedded in office boilerplate. The test distractors are disjoint sentence pools. One of the four kinds (office boilerplate) is the same kind of text as the training filler, and it is the only kind used in the offset, straddling, length and competing-evidence experiments, which carry the headline numbers. The distractor-type experiment adds the other three kinds (Section 5) and is the nearest we have to an out-of-family check for the filler; it keeps the evidence, generator and schemas fixed. Not tested are another generator, naturally long documents, other question schemas and non-English text. Every gain reported for v1_long is therefore an in-family result. The two fine-tunes also differ in training `max_len` and in the number of training rows, so the gain is the effect of this recipe and not of long-input exposure in the abstract. A control that adds the same number of short copies was not run.

### 3.4 Statistics

Accuracy is reported with Wilson 95% intervals, treating requests as independent, which overstates precision where one item appears at several positions. For differences between checkpoints we use paired comparisons on identical requests, with an exact McNemar test and a 95% bootstrap interval that resamples *items* and not requests. No correction for multiple comparisons is applied. About 30 paired differences are printed (15 between checkpoints in Table 14, 14 strategy-versus-default rows in Table 15 and the scan against the 2,048 window), and the strategy ranking was read off the same 80 items it is reported on. The headline differences have item-level intervals far from zero, and the near-zero ones are called noise. The paper is a descriptive study, and p-values flag large paired differences without deciding hypotheses. There is one training seed per checkpoint, one stream seed per item and one run of each timing, so differences of a few points between checkpoints should not be over-read.

## 4 Evidence-position curves

### 4.1 The default window, by offset

Figure 4 shows the main measurement. Each document has 1,800 tokens (about 1,799 ± 2 after tokenization of the spliced text), the evidence starts at a grid offset, and the same document is read with four `max_len` settings. Wherever the evidence lies *beyond* the end of the window, the answer is unrelated to it by construction, so these requests show how the models behave when the evidence is lost. Pooled over positions and settings, the released checkpoint answers correctly 26.4% [24.4-28.5] when the evidence is wholly cut (n=1,770) and 38.5% [37.2-39.8] when it is wholly inside (n=5,307); Table 3 gives the same split for the other checkpoints.

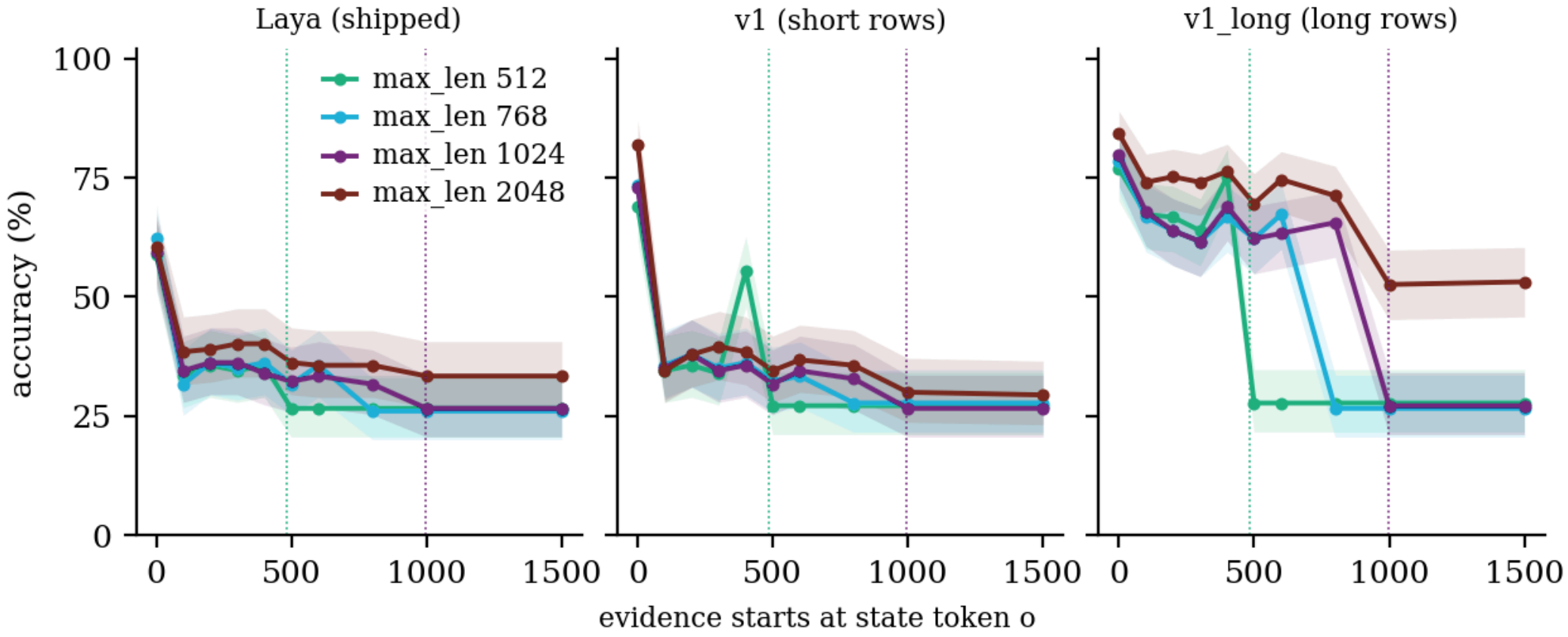

**Figure 4.** Accuracy against the absolute offset o of the evidence in a 1,800-token document, one panel per checkpoint, one line per `max_len` (bands: Wilson 95%). Dotted verticals: end of the window at 512 and 1024 for a 3-way question.

**Table 3.** Accuracy by what the window contained, pooled over the four `max_len` settings and all positions. Partly visible: some but not all evidence tokens inside the window (rare here, see the straddling experiment). Last columns: mean top probability.

| Checkpoint | Evidence cut | Partly visible | Fully visible | Conf. (cut) | Conf. (visible) |
|---|---|---|---|---|---|
| shipped | 26.4% [24.4-28.5] | 66.7% [20.8-93.9] | 38.5% [37.2-39.8] | 0.57 | 0.57 |
| v1 | 27.2% [25.2-29.3] | 100.0% [43.8-100.0] | 40.8% [39.5-42.1] | 0.67 | 0.74 |
| v1_long | 27.2% [25.2-29.4] | 100.0% [43.8-100.0] | 68.4% [67.2-69.7] | 0.82 | 0.87 |

Three features of the curves matter for the rest of the paper. The first is that in-window accuracy is not constant. With the whole 1,800 tokens visible (`max_len` 2048) the released checkpoint scores 60.5% when the evidence opens the state (o = 0) and 38.4% when only 100 tokens of boilerplate precede it. For the short-row fine-tune the figures are 81.9% and 34.5%, and for the long-row fine-tune 84.2% and 74.0%. The model does not search the text for the decisive passage. The numbers fit a model that judges the whole document it reads, so that a document that is mostly boilerplate reads as unremarkable; we did not test this reading directly. Only the checkpoint trained on long documents keeps most of its accuracy (Section 7 quantifies this).

The second feature is the cut itself, where accuracy falls to the evidence-free level and stays there. For the long-row checkpoint at `max_len` 512 accuracy goes from 75.1% at o = 400 to 27.7% at o = 500, and the steps for 768 and 1024 sit later, as the room dictates (Figure 5). For the other two checkpoints the step is small because only 12 points (released checkpoint) separate reading from not reading, so the in-window level is their main loss and the cut is a smaller one.

The third is that a long window is not read well beyond the training length. The long-row checkpoint was trained at 1,024 tokens. At `max_len` 2048 its accuracy is 74% on average for evidence at o = 100 to 800 but 53% at o = 1000 and 53% at o = 1500, although nothing is cut.

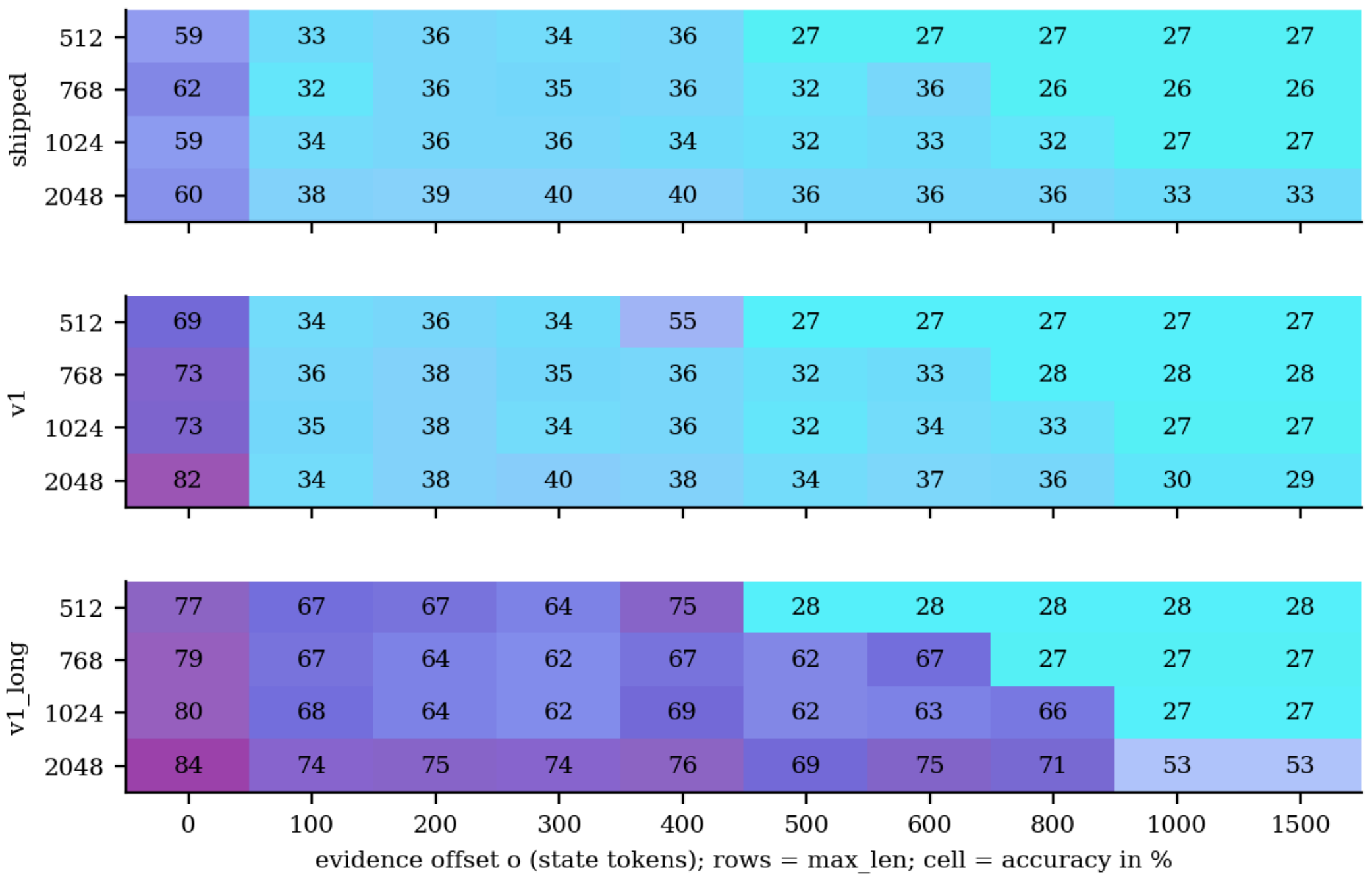

**Figure 5.** The same data as a grid: rows are `max_len`, columns evidence offsets, cells accuracy in %, for each checkpoint.

The evidence is only 20-60 tokens long and the grid spacing is 100, so the grid brackets the end of the window but does not resolve it. Table 4 compares the last grid offset that leaves the sentiment evidence wholly inside the window with the first one wholly beyond it; the straddling experiment below resolves the transition token by token.

**Table 4.** Accuracy on the sentiment question just before and just after the end of the window (room in tokens), by checkpoint and `max_len`. Offsets are the nearest grid points with the evidence wholly inside (o + 60 ≤ room) and wholly beyond (o ≥ room).

| Checkpoint | max_len | Room | o inside | Accuracy inside | o beyond | Accuracy beyond |
|---|---|---|---|---|---|---|
| shipped | 512 | 485 | 400 | 41.7% [27.1-57.8] | 500 | 19.4% [9.8-35.0] |
| shipped | 1024 | 997 | 800 | 38.9% [24.8-55.1] | 1000 | 27.8% [15.8-44.0] |
| v1 | 512 | 485 | 400 | 44.4% [29.5-60.4] | 500 | 30.6% [18.0-46.9] |
| v1 | 1024 | 997 | 800 | 36.1% [22.5-52.4] | 1000 | 30.6% [18.0-46.9] |
| v1_long | 512 | 485 | 400 | 66.7% [50.3-79.8] | 500 | 30.6% [18.0-46.9] |
| v1_long | 1024 | 997 | 800 | 52.8% [37.0-68.0] | 1000 | 30.6% [18.0-46.9] |

## 4.2 Position as a fraction of the document

Reading the evidence offset as an absolute number of tokens answers the question "where is the window". A user with documents of unpredictable length instead wants to know whether the beginning, the middle or the end of a document is read best. Figure 6 places the evidence at five relative positions in documents of about 600 and 1,300 tokens. For the released checkpoint with a window large enough to hold the 600-token document (`max_len` 1024) accuracy is 57.6% at the start, 35.0% in the middle and 41.8% at the end, with no cut involved. For the 1,300-token document at `max_len` 512 the end of the document is lost: accuracy for evidence at the end is 26.6%. The short-row fine-tune, with the same documents, scores 68% at the start, 36% in the middle and 60% at the end (177 requests per cell). In both, the middle of a document that is fully read is the weakest place and the start the strongest. This has the same ordering as the position effect reported for decoder models by Liu et al. [1] but is measured here on small cells. With intervals of about seven points per cell, the end-versus-middle gap is clear for the fine-tune and within noise for the released checkpoint. We make no claim beyond this document length.

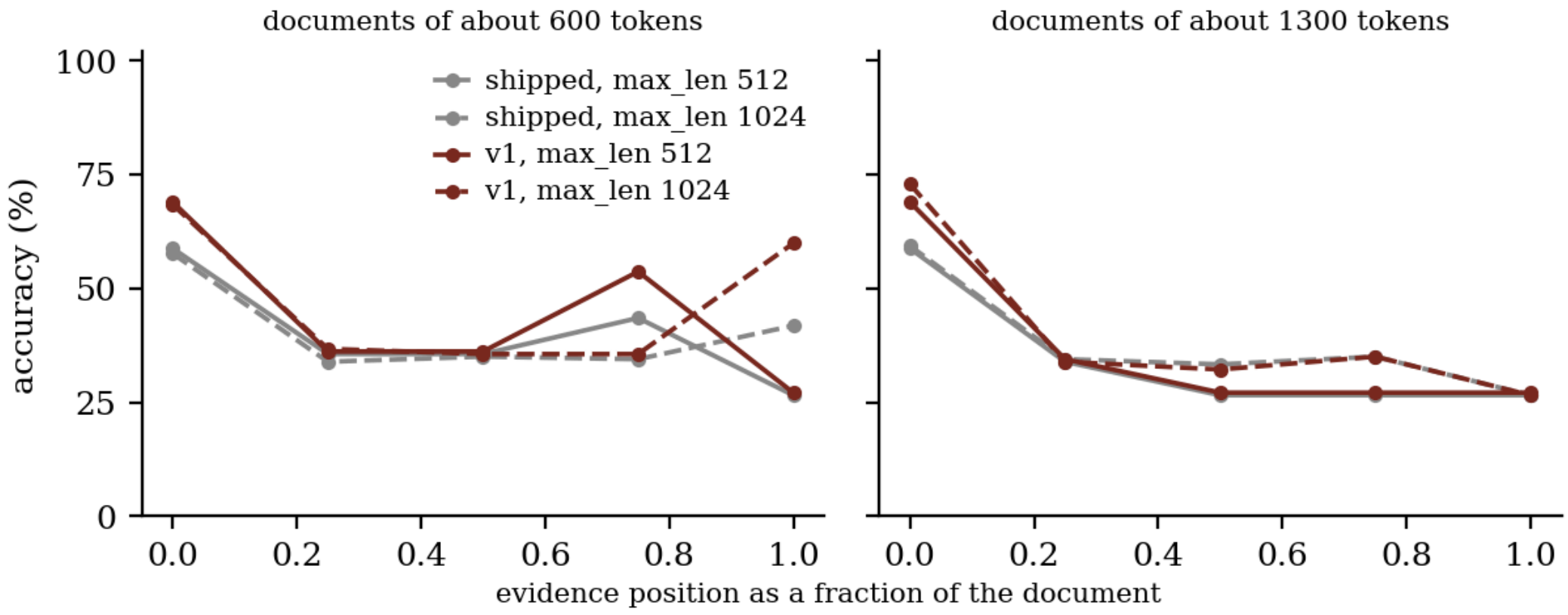

**Figure 6.** Accuracy against the evidence position as a fraction of the document, for documents of about 600 and 1,300 tokens (the window ends before the document does only for the longer one, and only at 512), at `max_len` 512 (solid) and 1024 (dashed).

### 4.3 Per-question effects

The room at 512 differs between questions only by the head (Table 1), but the questions differ in how much the model needs the evidence at all. Where the evidence-free answer is already right for a sizeable share of items (a two-way question, or a skewed one), a cut costs less accuracy. Figure 7 gives the per-question curves; the accuracy with the evidence inside versus cut is, for the released checkpoint, sentiment 47% against 21%; hardware 35% against 24%; urgency 47% against 42%; topic-12 16% against 4%; dept/churn 49% against 42%.

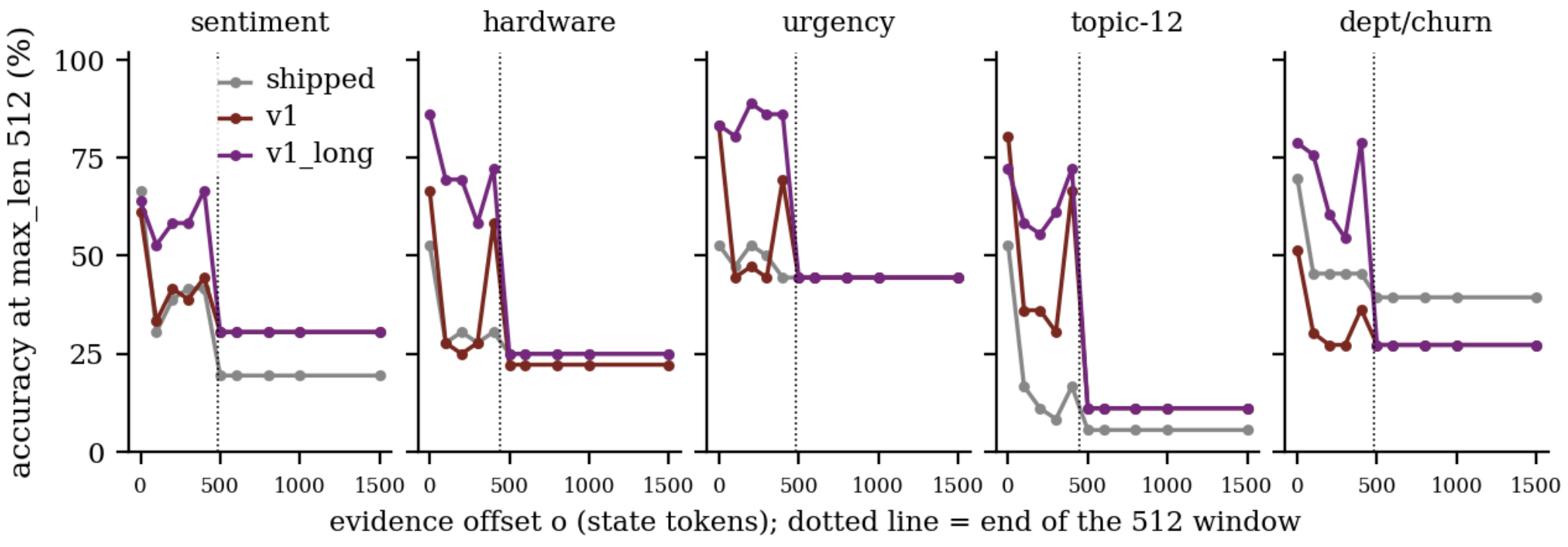

**Figure 7.** Per-question offset curves at `max_len` 512 for the three checkpoints; the dotted vertical line marks the end of the window for that question.

## 5 Cut tokens, distractors and competing evidence

### 5.1 Evidence that straddles the cut

If the transition at the end of the window is as sharp as the structure suggests, accuracy should follow the number of evidence tokens inside the window. We place the evidence so that a fraction v of its tokens lies before the end of the window (v = 0 means it begins exactly at the cut, v = 1 that it ends exactly at it). Figure 8 and Table 5 give the result.

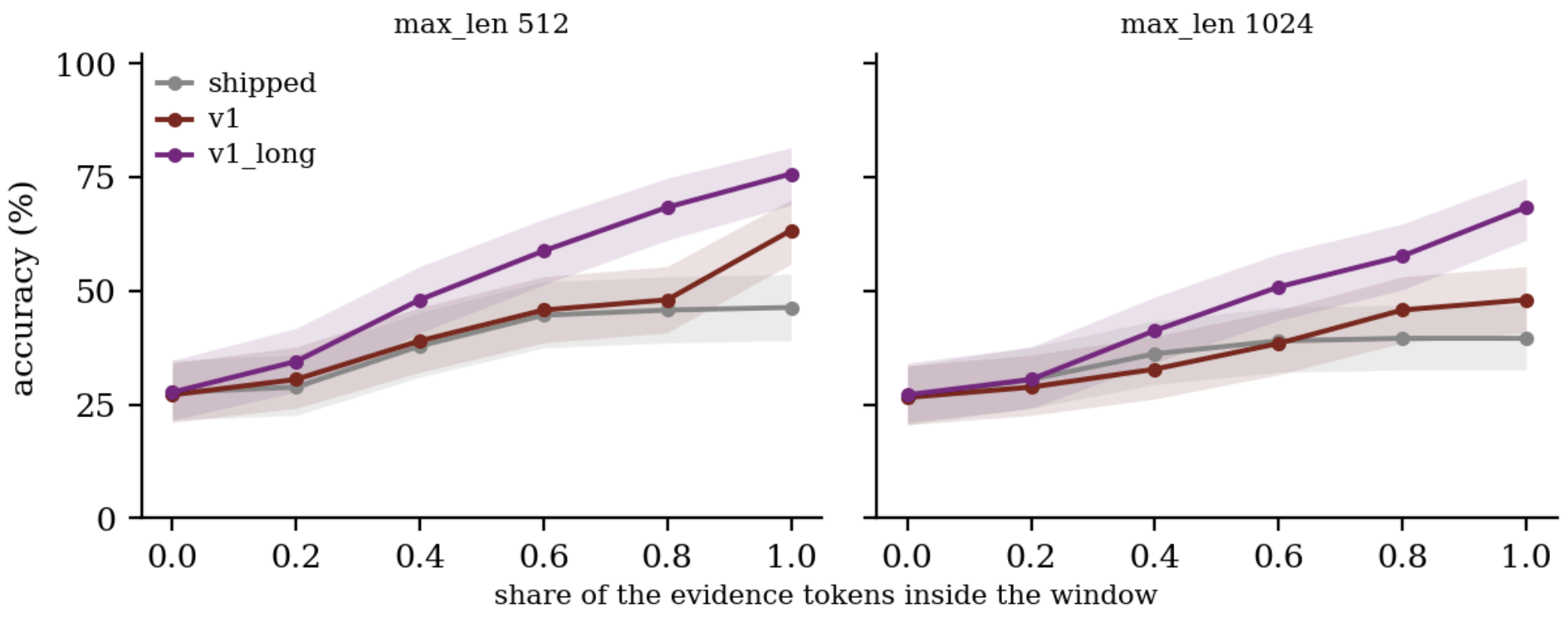

**Figure 8.** Evidence that straddles the end of the window: the share of its tokens that lie inside the window is varied from 0 to 100%, at `max_len` 512 and 1024 (Wilson bands).

**Table 5.** Accuracy by the share of the evidence inside the window at `max_len` 512 (the 1024 rows are in Figure 8).

| Checkpoint | max_len | 0% | 20% | 40% | 60% | 80% | 100% |
|---|---|---|---|---|---|---|---|
| shipped | 512 | 28% [22-35] | 29% [23-36] | 38% [31-45] | 45% [38-52] | 46% [39-53] | 46% [39-54] |
| v1 | 512 | 27% [21-34] | 31% [24-38] | 39% [32-46] | 46% [39-53] | 48% [41-55] | 63% [56-70] |
| v1_long | 512 | 28% [22-35] | 34% [28-42] | 48% [41-55] | 59% [51-66] | 68% [61-75] | 76% [69-81] |

The transition is gradual in the share of evidence tokens, even though it is sharp in position. For the long-row checkpoint at `max_len` 512 accuracy rises from 28% (no token visible) through 48% (40%) to 76% (all tokens), and the released checkpoint reaches 45% at 60% and 46% at 100%. Evidence of 20 to 60 tokens is partly informative when half of it is read, so the width of the cliff in position is about the length of the evidence, as expected (E1). A decisive passage of several hundred tokens would blur the cliff over that many tokens.

## 5.2 Tokens cut and document length

The straddling experiment fixes the document length at 1,800 tokens. Here the length varies and the evidence is placed either 100 tokens from the start or at the very end, so the number of tokens cut grows with the length. The head placement keeps the evidence inside the window and isolates a pure length effect. In the tail placement the evidence is lost as soon as the document exceeds the room, so length becomes evidence loss.

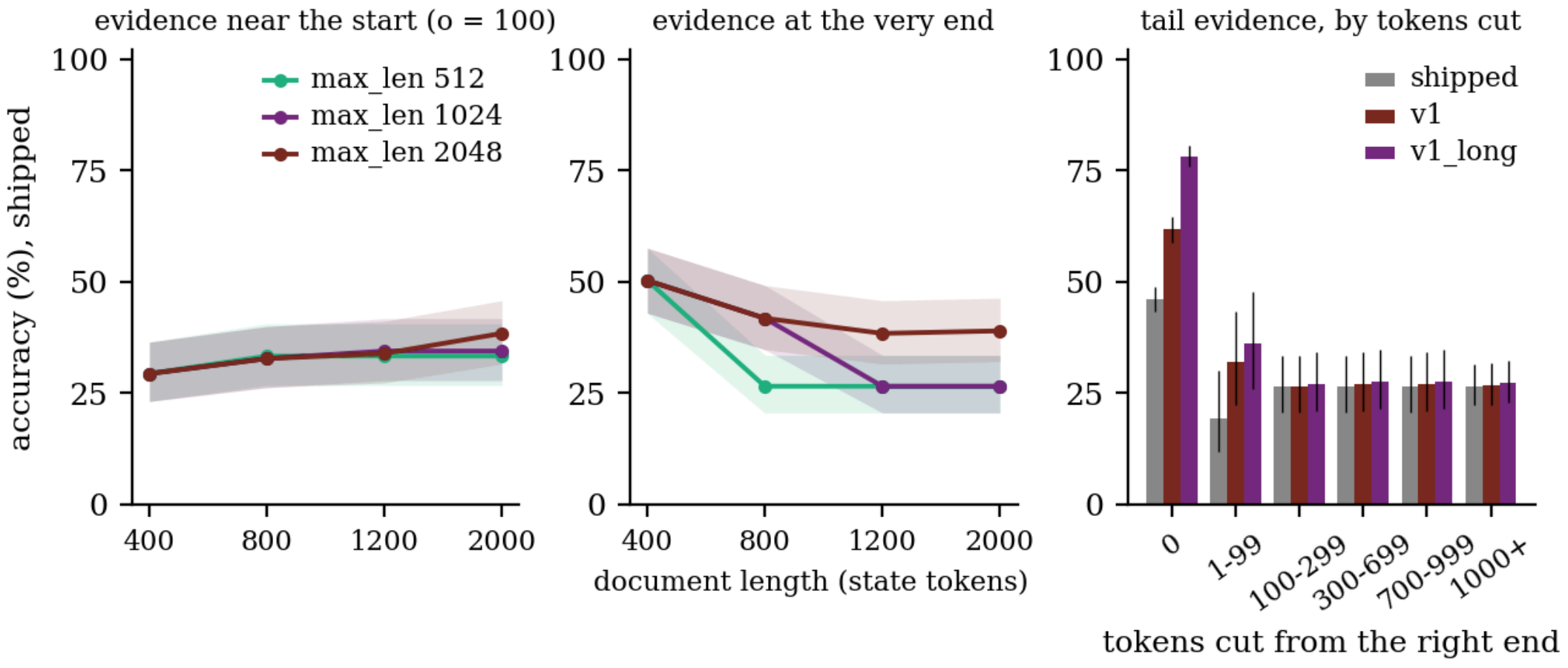

**Figure 9.** Document length swept with the evidence either near the start (left) or at the very end (middle); right: accuracy of the tail placement by the number of tokens cut from the right.

**Table 6.** Accuracy by document length (state tokens) at `max_len` 2048 for evidence near the start ('head', o = 100) or at the very end ('tail'); the values at 512 are in the text and Figure 9.

| Checkpoint | Evidence | max_len | 400 | 800 | 1200 | 2000 |
|---|---|---|---|---|---|---|
| shipped | head | 2048 | 29% [23-36] | 33% [26-40] | 34% [27-41] | 38% [32-46] |
| shipped | tail | 2048 | 50% [43-58] | 42% [35-49] | 38% [32-46] | 39% [32-46] |
| v1 | head | 2048 | 36% [29-43] | 34% [27-41] | 35% [28-42] | 36% [29-43] |

| Checkpoint | Evidence | max_len | 400 | 800 | 1200 | 2000 |
|---|---|---|---|---|---|---|
| v1 | tail | 2048 | 70% [63-76] | 60% [53-67] | 50% [42-57] | 40% [33-47] |
| v1_long | head | 2048 | 67% [60-74] | 66% [58-72] | 69% [62-75] | 76% [69-81] |
| v1_long | tail | 2048 | 83% [77-88] | 75% [68-81] | 72% [65-78] | 59% [52-66] |

Text after the evidence costs nothing as it grows. With the evidence near the start, accuracy at `max_len` 512 is flat from 400 to 2,000 tokens (29% → 33% → 33% → 33% for the released checkpoint, 36% → 34% → 34% → 34% for the short-row fine-tune, 67% → 67% → 67% → 67% for the long-row one), although up to 1,600 tokens are cut. Text before the evidence costs a lot, and the cost grows with its length while the evidence is still read. With the evidence last, the long-row checkpoint at 2048 declines (83% → 75% → 72% → 59%) as the document grows from 400 to 2,000 tokens. The others decline too (50% → 42% → 38% → 39% for the released and 70% → 60% → 50% → 40% for the short-row checkpoint). At 512 and 1024 the tail evidence is lost as soon as the document outgrows the room (83% → 28% → 28% → 28% and 83% → 75% → 27% → 27% for the long-row checkpoint). The last panel of Figure 9 bins the same requests by the number of tokens cut, and a few tokens are enough to lose the evidence. At 2048 the 2,000-token documents no longer fit next to the heavier question heads, so part of the last drop is again a cut.

## 5.3 Distractor length and type

With the window at 2,048 nothing is cut, so any loss is caused by the text itself. We add 0, 100, 300 or 800 distractor tokens before, around or after the evidence and vary the *kind* of text. The four kinds are office boilerplate (neutral with respect to every question), encyclopedic sentences, rows from the other tasks (text that looks like the input of a decision model, with sentiment or faults of its own and so similar in form to the evidence) and, for sentiment only, neutral-sentiment rows of the same task (on topic but pointing the other way).

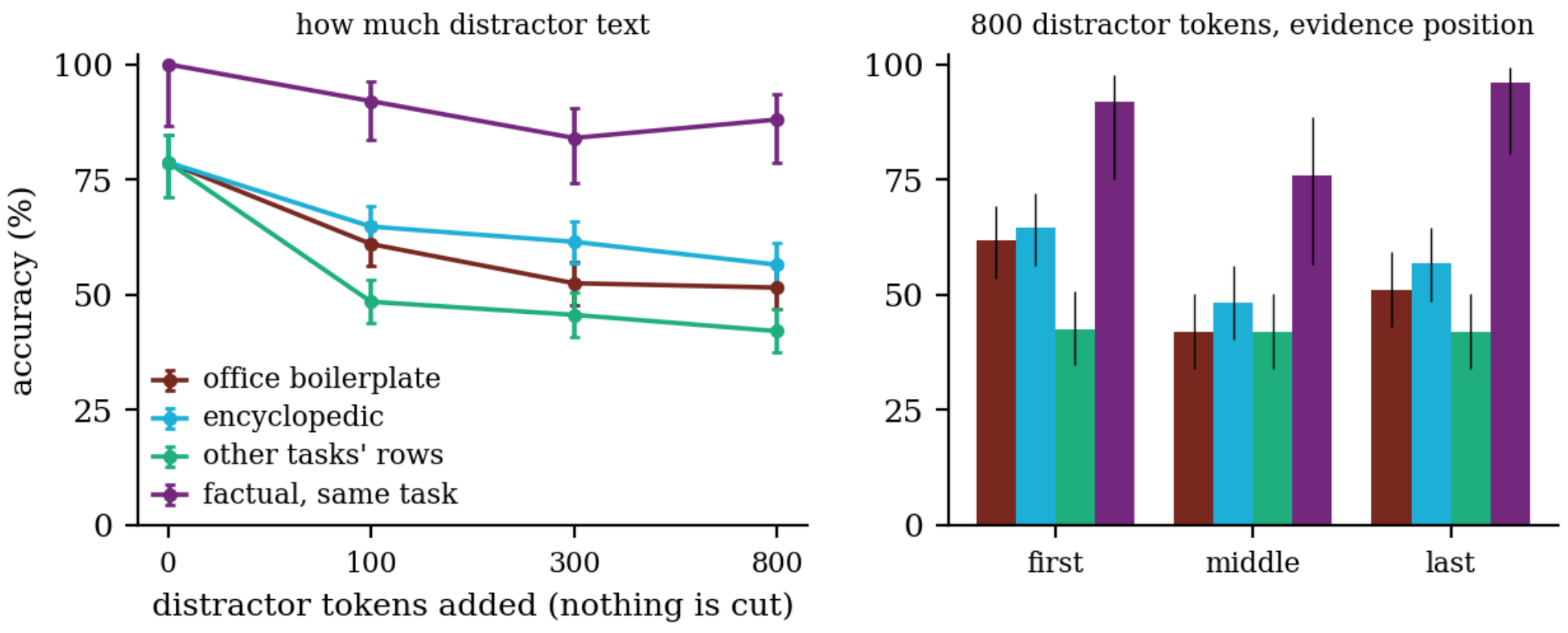

**Figure 10.** No cut (`max_len` 2048): accuracy of the released checkpoint against the amount of distractor text added to the evidence (left, evidence position averaged) and against the evidence position among 800 distractor tokens (right), for four kinds of distractor text.

**Table 7.** Accuracy by amount of distractor text added (tokens; position-averaged, no truncation) and mean confidence at 800 tokens. The 0-token column is the evidence without distractors; for the last kind it is restricted to the sentiment items that kind uses.

| Checkpoint | Distractor text | 0 | 100 | 300 | 800 | Mean conf. at 800 |
|---|---|---|---|---|---|---|
| shipped | office boilerplate | 79% [71-85] | 61% [56-66] | 52% [48-57] | 52% [47-56] | 0.55 |
| shipped | encyclopedic | 79% [71-85] | 65% [60-69] | 61% [57-66] | 57% [52-61] | 0.59 |
| shipped | other tasks' rows (similar form) | 79% [71-85] | 48% [44-53] | 46% [41-50] | 42% [37-47] | 0.64 |
| shipped | same-task neutral rows (on topic, contradicting) | 100% [87-100] | 92% [84-96] | 84% [74-91] | 88% [79-94] | 0.71 |
| v1 | office boilerplate | 95% [90-98] | 62% [58-67] | 54% [50-59] | 55% [50-59] | 0.86 |

| Checkpoint | Distractor text | 0 | 100 | 300 | 800 | Mean conf. at 800 |
|---|---|---|---|---|---|---|
| v1 | encyclopedic | 95% [90-98] | 67% [62-71] | 60% [56-65] | 62% [57-66] | 0.87 |
| v1 | other tasks' rows (similar form) | 95% [90-98] | 64% [59-68] | 52% [47-56] | 44% [40-49] | 0.91 |
| v1 | same-task neutral rows (on topic, contradicting) | 100% [87-100] | 52% [41-63] | 45% [35-57] | 53% [42-64] | 0.77 |
| v1_long | office boilerplate | 94% [89-97] | 77% [73-81] | 76% [71-79] | 76% [72-80] | 0.92 |
| v1_long | encyclopedic | 94% [89-97] | 77% [73-81] | 72% [67-76] | 80% [76-83] | 0.92 |
| v1_long | other tasks' rows (similar form) | 94% [89-97] | 61% [57-66] | 48% [44-53] | 44% [40-49] | 0.89 |
| v1_long | same-task neutral rows (on topic, contradicting) | 100% [87-100] | 68% [57-77] | 64% [53-74] | 73% [62-82] | 0.85 |

For the released checkpoint, office boilerplate costs 18 points at 100 tokens and 27 points at 800 (Table 7). The loss is almost complete after the first 100 to 300 tokens and does not keep growing. Encyclopedic sentences cost somewhat less (57% at 800 tokens), and rows from other tasks cost more (42%), as one would expect when the distractors look like inputs of the model. Position matters inside the window. With 800 boilerplate tokens the released checkpoint scores 62% when the evidence is first, 42% in the middle and 51% last, a start-and-end advantage over the middle that resembles the pattern reported for decoder models [1]. For the sentiment items whose distractors are neutral-sentiment rows of the same task the two models react differently. The released checkpoint falls to 84% at 300 tokens (baseline 100%) and the short-row fine-tune to 45% (baseline 100%), so the fine-tune is pulled away from the positive or negative evidence towards the neutral distractors. The long-row checkpoint, run on the same documents, gives 76% for office boilerplate, 80% for encyclopedic, 44% for other tasks' rows (similar form), 73% for same-task neutral rows (on topic, contradicting) at 800 tokens.

### 5.4 What does the model answer when there is no evidence?

Documents from which the evidence has been removed show what the answer looks like once the decisive text is lost, which is what a cut produces. The mean confidence of these answers is 0.57 for 800 tokens of boilerplate at 2,048 and 0.57 at 512, to be compared with the confidence of ordinary requests in Section 6. The answers are concentrated on one option. For the 3-way sentiment question the most frequent answer is given in 90.7% of the evidence-free documents by the released checkpoint and in 100% by the short-row fine-tune. Accuracy therefore equals the share of items whose gold label happens to be that option (Table 8). This is not a fixed chance level. Pooled over the five questions the cut requests score 26.4% for the released checkpoint, against 27.0% for a uniform guess and 35.6% for always giving the commonest label. Per question it falls below the uniform rate for sentiment (21% against 33%) and topic (4% against 8%) and above it for urgency (42% against 33%). Each checkpoint defaults to a different answer, so the pooled values of the three checkpoints (26.4%, 27.2%, 27.2%) agree more closely than their per-question values do. The fine-tuned checkpoints are the more confident ones when nothing is there to read: mean confidence 0.67 (short-row) and 0.83 (long-row) at 800 tokens, close to what they show for requests that do contain evidence (Section 6).

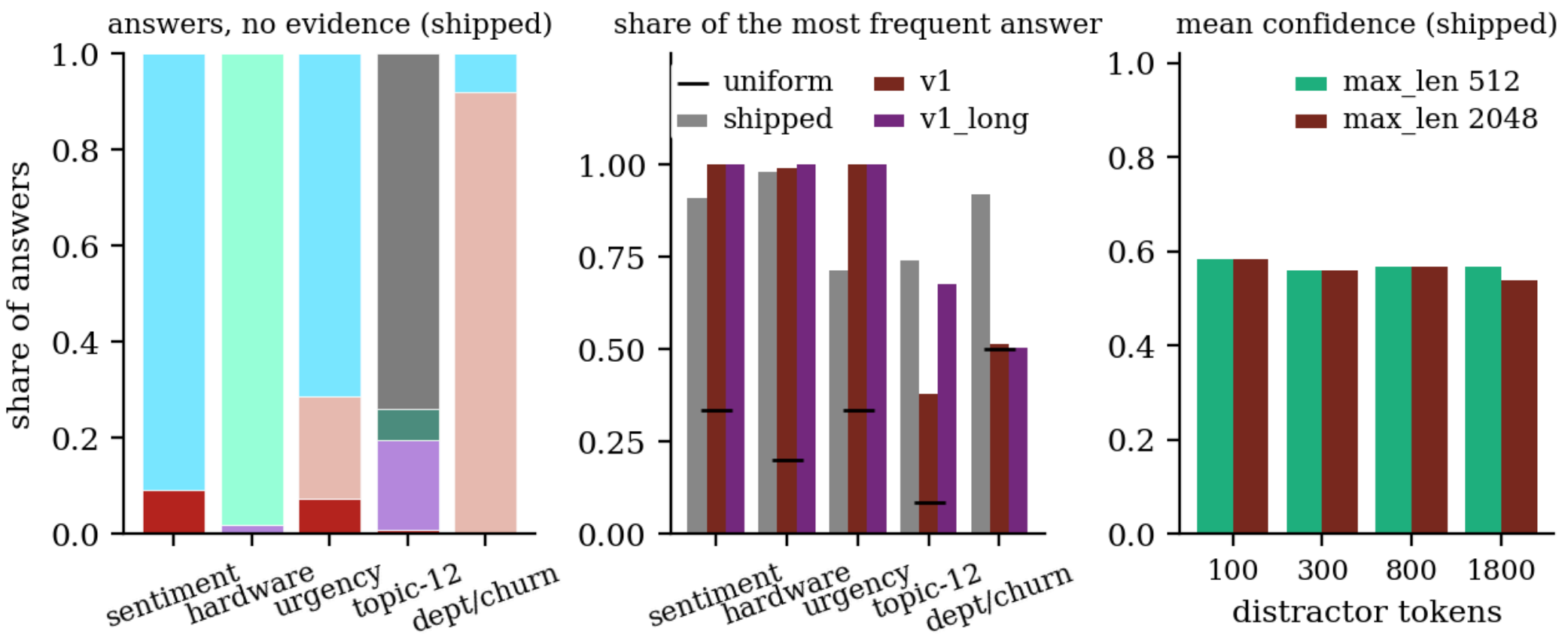

**Figure 11.** Documents without any evidence. Left: which options the released checkpoint answers (neutral boilerplate, 300-1,800 tokens, `max_len` 2048; one colour per option). Middle: share of the single most frequent answer, by checkpoint. Right: mean confidence of the answers as the amount of distractor text grows.

**Table 8.** Evidence-free documents (released checkpoint, neutral boilerplate of 300-1,800 tokens, `max_len` 2048): the most frequent gold label of the question, the most frequent answer, and the accuracy against the items' gold labels.

| Question | Majority gold label | Most frequent answer | Accuracy vs gold |
|---|---|---|---|
| sentiment, 3-way | 41.7% | 90.7% | 25% [18-34] |
| hardware failure, 5-way | 25.0% | 98.1% | 25% [18-34] |
| urgency, 3 levels | 47.2% | 71.3% | 39% [30-48] |
| topic, 12-way | 16.7% | 74.1% | 3% [1-8] |
| department / churn | 48.5% | 91.9% | 42% [33-52] |

## 5.5 Competing evidence

A real long document often says more than one thing. We join two rows with different labels (gap 0) or separate them by 100, 400 or 900 tokens of neutral boilerplate. We then count which one the answer follows. When the second piece falls beyond the window the first can only win or lose to a third answer. When both are visible, the comparison shows which end of the input the model prefers. With the pieces adjacent the released checkpoint follows the first in 32% and the second in 68% of the sentiment pairs. At a gap of 900 tokens and `max_len` 512 the second piece is cut and the first wins 84% of the time, compared with 71% when both pieces are read.

**Table 9.** Which evidence wins: share of answers following the first / second piece of evidence (released checkpoint, 80 pairs per question, all other answers belong to neither).

| Question | max_len | gap 0 | gap 100 | gap 400 | gap 900 |
|---|---|---|---|---|---|
| sentiment | 512 | 32% / 68% | 44% / 46% | 48% / 52% | 84% / 9% |
| sentiment | 1024 | 32% / 68% | 44% / 46% | 48% / 52% | 71% / 28% |
| sentiment | 2048 | 32% / 68% | 44% / 46% | 48% / 52% | 71% / 28% |
| hardware | 512 | 21% / 44% | 18% / 41% | 22% / 30% | 39% / 21% |
| hardware | 1024 | 21% / 44% | 18% / 41% | 18% / 30% | 20% / 31% |
| hardware | 2048 | 21% / 44% | 18% / 41% | 18% / 30% | 20% / 31% |

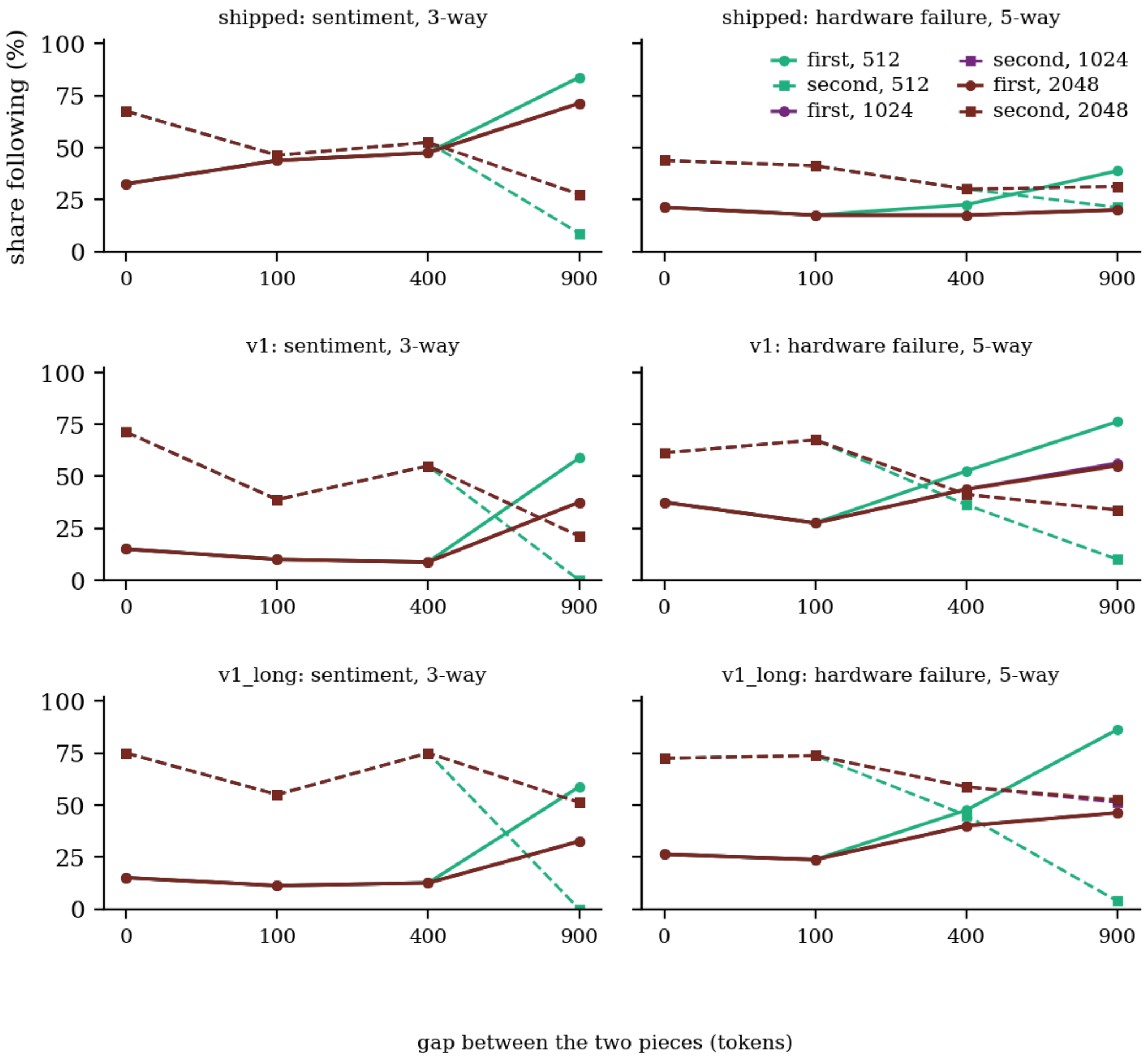

**Figure 12.** Two pieces of evidence with different labels separated by neutral text: share of answers that follow the first (solid) or the second (dashed) one, by gap and `max_len` (line colour, see legend; the lines coincide at small gaps), for sentiment and hardware failure.

The preference is for the *later* piece only when the two are adjacent. With the pieces adjacent, the share of answers following the first piece against the second is 32% / 68%; 15% / 71%; 15% / 75% for sentiment and 21% / 44%; 38% / 61%; 26% / 72% for hardware failure (by checkpoint in the order shipped, v1, v1_long). For the adjacent case this contradicts the expectation E4 that the first piece wins by default. With 100 or 400 neutral tokens between them the released checkpoint follows the two sentiment pieces about equally (44% / 46% and 48% / 52%; 80 pairs per cell, so these gaps are within noise) and still favours the later hardware piece. The two fine-tuned checkpoints follow the later sentiment piece at every gap up to 400 tokens (first / second 9% / 55% for v1 and 12% / 75% for v1_long at 400). We call the adjacent case a recency preference and do not test why it arises. The gap-900 column is dominated by the window: at 512 the second piece is cut for sentiment and the first piece then wins by default.

## 6 Does the output reveal the cut?

A downstream system that gates on the model's probability would be protected from the cut if confidence dropped when evidence is missing. Figure 13 compares the probability of the top option for requests whose evidence was entirely cut with those whose evidence was entirely inside. These come from the offset experiment, where every request has the same 1,800-token length and the two groups differ only in `max_len` and offset. For the released checkpoint the mean top probability is 0.57 when the evidence is cut and 0.57 when it is read, and the AUROC of probability as a detector is 0.50, which is chance. The long-row checkpoint is more confident overall (0.82 against 0.87) and reaches an AUROC of only 0.55.

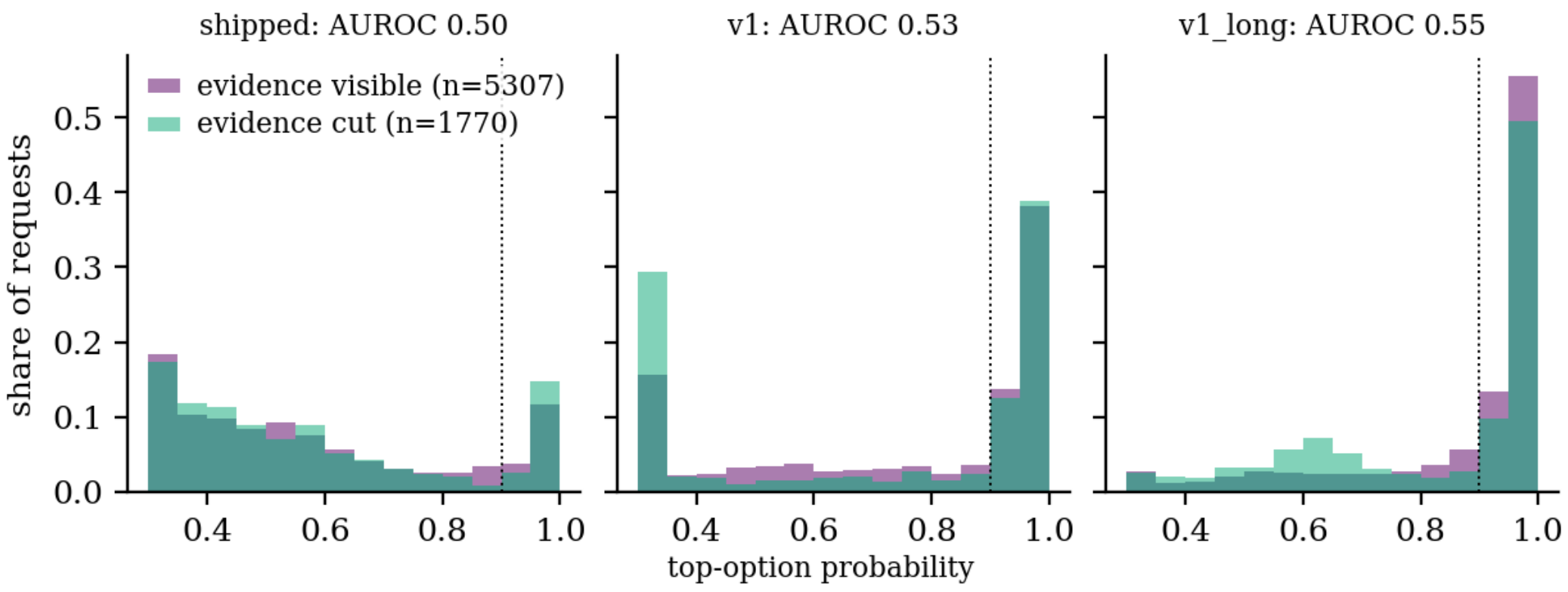

**Figure 13.** Top-option probability of requests whose evidence was wholly cut and wholly inside the window (legend in each panel), by checkpoint (offset experiment, all `max_len` settings pooled). Dotted line: a 0.9 gate. The AUROC in each title is the chance that a cut request is less confident than a visible one.

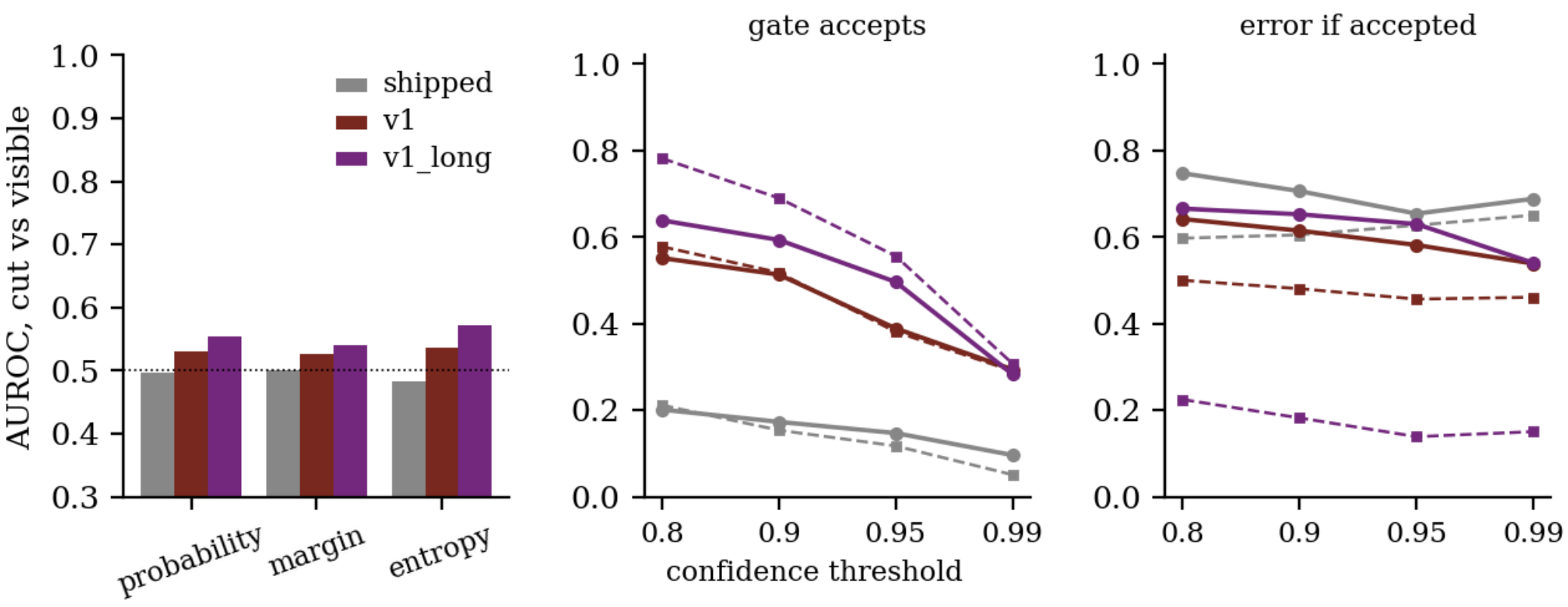

**Figure 14.** Left: AUROC of three confidence scores for telling cut from visible requests (0.5 = no information). Middle and right: share of requests a confidence gate accepts and the error rate among those it accepts, for cut (solid) and visible (dashed) requests.

**Table 10.** Can confidence tell that the evidence was cut? AUROC for separating cut from fully visible requests (0.5 = chance), overall and per question, and mean top probability in each group.

| Checkpoint | Prob. | Margin | Entropy | Mean conf. cut | Mean conf. visible | sentiment | hardware | urgency | topic-12 | dept/churn |
|---|---|---|---|---|---|---|---|---|---|---|
| shipped | 0.498 | 0.501 | 0.484 | 0.568 | 0.567 | 0.49 | 0.41 | 0.55 | 0.50 | 0.43 |
| v1 | 0.530 | 0.527 | 0.536 | 0.674 | 0.740 | 0.34 | 0.77 | 0.41 | 0.71 | 0.46 |
| v1_long | 0.554 | 0.540 | 0.571 | 0.824 | 0.875 | 0.24 | 0.77 | 0.53 | 0.65 | 0.54 |

Table 11 turns this into the operating point that matters in practice. At a threshold of 0.9, the gate accepts 17.3% of the cut requests and 15.4% of the visible ones for the released checkpoint, and 70.6% of the accepted cut requests are wrong. The reason is structural: the probabilities are computed from a shorter document, and nothing in them knows what is missing. For the long-row checkpoint the same gate is informative for requests that were read (error 18% among accepted visible requests) and misleading for cut ones (error 65%). A gate that works on short test inputs therefore gives false assurance on long ones. A confidence gate sees the cut only to the extent that the shorter document happens to look less decisive.

**Table 11.** What a confidence gate does with cut and visible requests. ‘Accepted’ = top probability at or above the threshold; error = share of accepted requests that are wrong.

| Checkpoint | Threshold | Cut: accepted | Cut: error among accepted | n accepted (cut) | Visible: accepted | Visible: error among accepted |
|---|---|---|---|---|---|---|
| shipped | 0.90 | 17.3% | 70.6% | 306 | 15.4% | 60.5% |
| shipped | 0.99 | 9.6% | 68.8% | 170 | 5.1% | 65.1% |
| v1 | 0.90 | 51.3% | 61.5% | 908 | 51.8% | 48.1% |
| v1 | 0.99 | 29.4% | 53.8% | 520 | 29.0% | 46.1% |

| Checkpoint | Threshold | Cut: accepted | Cut: error among accepted | n accepted (cut) | Visible: accepted | Visible: error among accepted |
|---|---|---|---|---|---|---|
| v1_long | 0.90 | 59.3% | 65.2% | 1050 | 69.0% | 18.2% |
| v1_long | 0.99 | 28.2% | 54.0% | 500 | 30.7% | 15.0% |

### 6.1 The truncation flag

The runtime exposes a flag, and Table 12 evaluates it as a warning for lost evidence. In the library source the flag is true whenever fewer state tokens are kept than the state has, so it depends on the length of the text and never on where the decisive part sits; the table applies the same rule to our requests. Its recall is perfect by construction, since a cut that hides evidence is a cut. Its precision is the share of flagged requests whose evidence really was cut: 33.3% for the released checkpoint in this design, because a long document with the evidence early is flagged too. That figure is set by the design and says nothing about real traffic: every 1,800-token document is longer than the room at three of the four window sizes. What carries over is the mechanism. The flag tells a deployer that *something* was cut and not that something *decisive* was cut, and a gate on it would reject 66.6% of flagged requests here whose evidence was perfectly readable. Unlike confidence it never misses a cut, so where it is the only signal available it is the better of the two. Where the position of decisive text is unknown, the safe response to a flag is to re-read with a larger window or a different strategy (Sections 8 and 10) and not to reject.

**Table 12.** The runtime's truncation flag as a warning. A request is flagged when the state is longer than the room. Recall: share of cut-evidence requests that are flagged. The other columns split the flagged requests by what the window contained; the last is the error rate of flagged requests whose evidence was nevertheless fully inside.

| Checkpoint | Flagged requests | Recall (cut) | Flagged: evidence cut | partly visible | fully visible | Error among flagged, evidence visible |
|---|---|---|---|---|---|---|
| shipped | 5,310 | 100.0% | 33.3% | 0.1% | 66.6% | 61.9% |
| v1 | 5,310 | 100.0% | 33.3% | 0.1% | 66.6% | 58.7% |
| v1_long | 5,310 | 100.0% | 33.3% | 0.1% | 66.6% | 32.6% |

## 7 Does long-input training add anything beyond a larger window?

Two levers can address the loss. A configuration change raises `max_len` and needs no retraining. A training change fine-tunes on long documents, as the third checkpoint was. They are not interchangeable. A larger window is necessary for evidence beyond the old one to be read at all, whatever the weights, while training can change *how well* a checkpoint reads what it is given. Table 13 puts both on one grid.

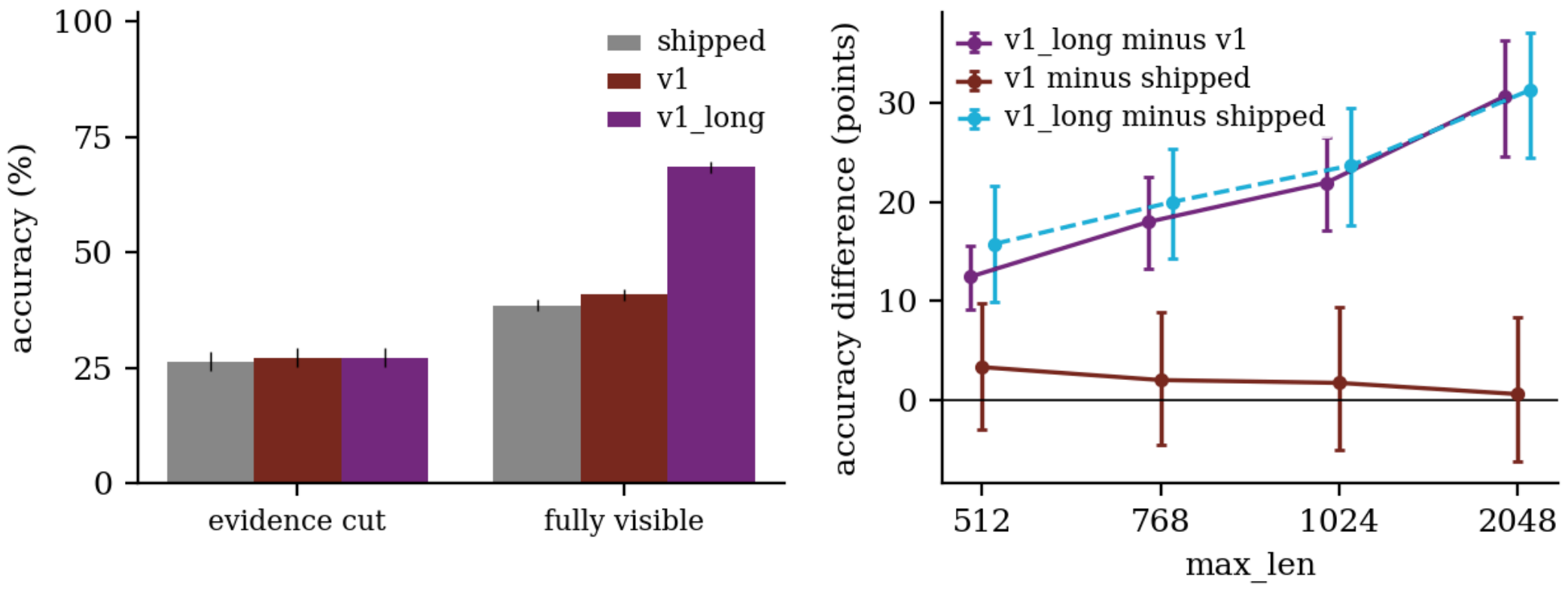

**Figure 15.** Left: accuracy when the evidence was cut or fully inside the window, for the three checkpoints (offset experiment, all settings pooled; the three partly visible requests are omitted; bars: Wilson 95%). Right: paired accuracy difference between checkpoints on identical requests at each `max_len` (bars: bootstrap interval over items).

**Table 13.** Accuracy pooled over all evidence positions, by checkpoint and `max_len`: moving down a column is the effect of training, moving right along a row the effect of the configuration change.

| Checkpoint | max_len 512 | max_len 768 | max_len 1024 | max_len 2048 |
|---|---|---|---|---|
| shipped | 33.1% [30.9-35.3] | 34.6% [32.5-36.9] | 35.0% [32.8-37.3] | 39.2% [37.0-41.5] |
| v1 | 36.4% [34.2-38.7] | 36.7% [34.5-38.9] | 36.8% [34.6-39.1] | 39.8% [37.6-42.1] |

| Checkpoint | max_len 512 | max_len 768 | max_len 1024 | max_len 2048 |
|---|---|---|---|---|
| v1_long | 48.8% [46.5-51.1] | 54.6% [52.3-56.9] | 58.7% [56.4-61.0] | 70.5% [68.3-72.5] |

**Table 14.** Paired comparisons between checkpoints on identical requests (accuracy difference in points, bootstrap 95% interval over items, exact McNemar p).

| Comparison | Subset | n | Acc. first | Acc. second | Difference [95%] | p |
|---|---|---|---|---|---|---|
| v1_long – v1 | all requests | 7,080 | 37.4% | 58.1% | +20.7 [+16.5, +25.2] | p<0.001 |
| v1_long – v1 | max_len 512 | 1,770 | 36.4% | 48.8% | +12.4 [+9.1, +15.5] | p<0.001 |
| v1_long – v1 | max_len 2048 | 1,770 | 39.8% | 70.5% | +30.6 [+24.6, +36.2] | p<0.001 |
| v1_long – v1 | evidence cut | 1,770 | 27.2% | 27.2% | +0.1 [-2.7, +2.7] | p=1.000 |
| v1_long – v1 | fully visible | 5,307 | 40.8% | 68.4% | +27.6 [+22.3, +33.1] | p<0.001 |
| v1 – shipped | all requests | 7,080 | 35.5% | 37.4% | +1.9 [-4.4, +9.1] | p=0.003 |
| v1 – shipped | evidence cut | 1,770 | 26.4% | 27.2% | +0.8 [-6.1, +7.1] | p=0.525 |
| v1 – shipped | fully visible | 5,307 | 38.5% | 40.8% | +2.3 [-4.2, +9.6] | p=0.004 |
| v1_long – shipped | all requests | 7,080 | 35.5% | 58.1% | +22.7 [+16.9, +28.0] | p<0.001 |
| v1_long – shipped | evidence cut | 1,770 | 26.4% | 27.2% | +0.8 [-6.5, +7.9] | p=0.504 |
| v1_long – shipped | fully visible | 5,307 | 38.5% | 68.4% | +29.9 [+23.9, +35.1] | p<0.001 |

Where the checkpoint cannot read distractor-laden text, the configuration change helps little. Moving the window from 512 to 2048 on identical requests changes accuracy by +6.2 points for the released checkpoint [+2.5, +10.1], +3.4 for the short-row fine-tune [+1.6, +5.5] and +21.6 for the long-row fine-tune [+17.8, +25.2]. Most of the gain is evidence that was cut and can now be read. For the first two checkpoints it is small because even readable evidence is answered at 38% and 41%.

Long-input training adds a large effect that is specific to visible evidence. Against the short-row fine-tune, the long-row checkpoint is +20.7 points better over all requests [+16.5, +25.2] and +27.6 points better when the evidence is fully visible [+22.3, +33.1], but +0.1 when it is cut [-2.7, +2.7]. Training does not let the model use what it cannot see. The gain grows with the window (+12.4 at 512, +21.9 at 1024, +30.6 at 2048), because at 512 a part of the evidence is cut for both checkpoints and at 2048 none is. The short-row fine-tune differs from the released checkpoint by +1.9 points [-4.4, +9.1]. The exact McNemar test on the 7,080 requests gives p=0.003, but the item-level interval includes zero, so we treat this as within noise because the same item appears in many requests.

For expectation E6, fine-tuning on long inputs helps only where the baseline cannot already use the longer window. Here the baselines cannot: at 2048 the long-row checkpoint is 31 points above the short-row one on identical requests, while the window change alone moved the short-row checkpoint by 3. Two cautions apply. First, the long training rows are embedded in office boilerplate, the same kind of text as the filler in the offset, straddling, length and competing-evidence experiments, so the headline gain is measured on the training distractor type. The distractor-type experiment (Section 5, `max_len` 2048, nothing cut) separates the kinds. Long-row minus short-row is +19.2 points [+15.2, +23.7] for office boilerplate, +13.2 [+9.9, +16.6] for encyclopedic sentences that the training rows never contained, +18.2 [+9.8, +27.6] for same-task neutral rows (sentiment items only) and -2.0 [-5.8, +1.3] for rows of other tasks. The gain therefore carries over to a filler it was not trained on when that filler is neutral, and not to filler that resembles the inputs of the model. Second, the gain fades beyond the training length (Section 4, third feature), so a longer window than the checkpoint was trained at is not free either.

## 8 Training-free alternatives to the right-hand cut

Fitting a long text into a 512-token window does not have to mean the right-hand cut. Without changing weights or configuration we can keep the end instead of the start, keep both ends, choose which sentences to keep, or read the document in overlapping windows and combine the answers. All variants below keep the question head and the window at 512 tokens (the last two rows use a larger window for comparison) and use the identical documents.

- *Tail only*: keep the last *room* tokens.

- *Head + tail*: keep the first quarter (or half) of the room and the rest from the end, a common heuristic in text-classification pipelines.
- *Lexical selection*: split the document into sentences and score each by the inverse-document-frequency weight of the words it shares with the question text and the option names. Keep the best-scoring sentences, in original order, until the room is full.
- *Window scan*: windows of 312 tokens with stride 156. For choice and score questions take the window with the highest top probability, and for yes/no questions the largest probability of the positive answer. This is the rule used by the library's own long-input method. We re-implemented it so that the windows can be batched and de-duplicated, and on 13 documents our version and the library call pick the same answer in 100.0% of the cases (Appendix E).

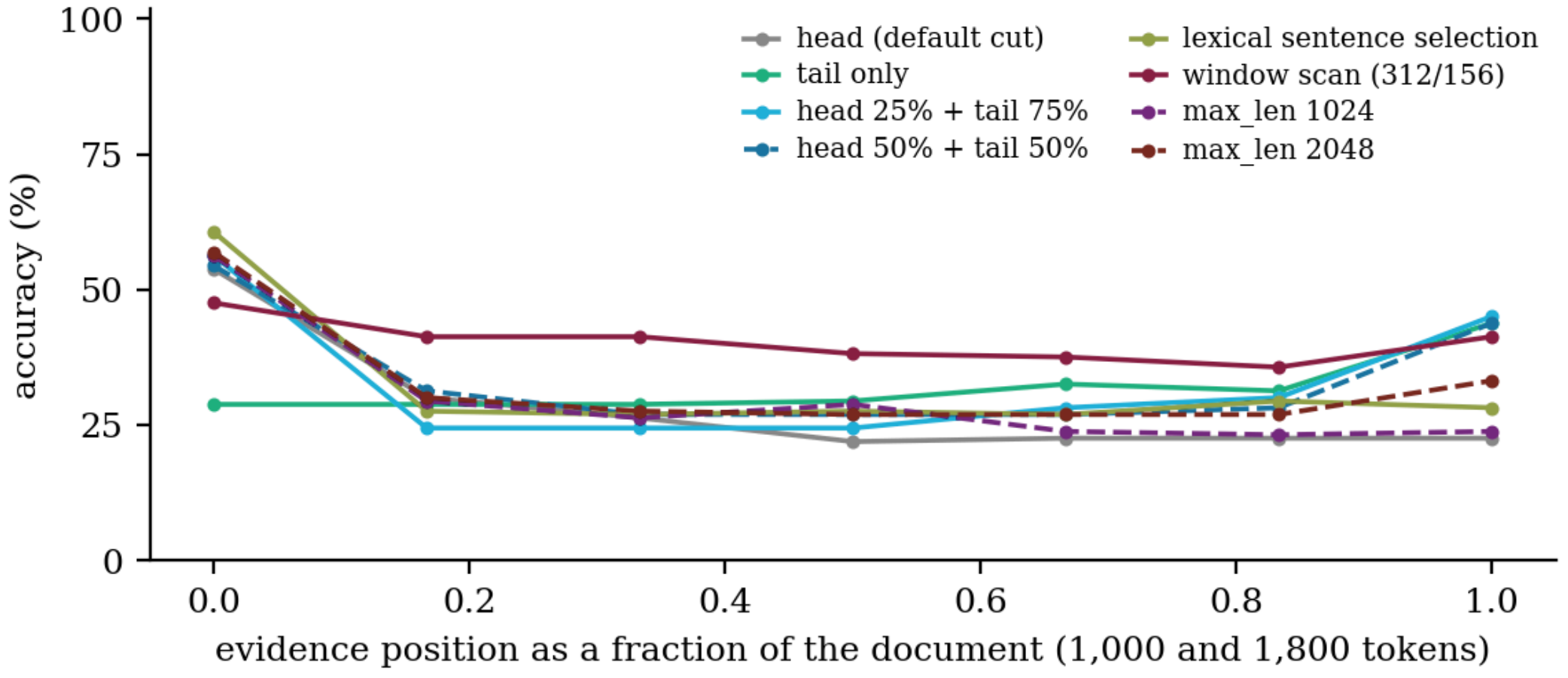

**Figure 16.** Training-free strategies at `max_len` 512 (and the two larger windows for reference) on documents of 1,000 and 1,800 tokens, by the evidence position as a fraction of the document (released checkpoint).

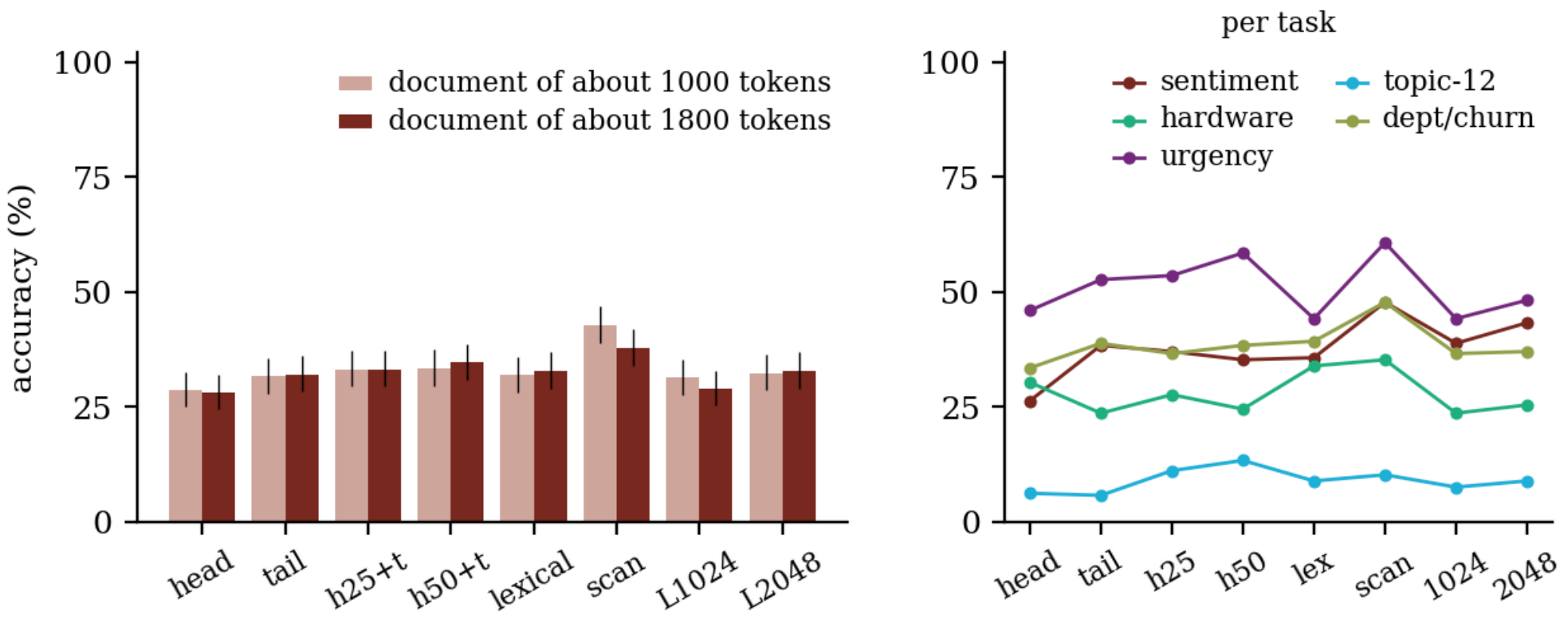

**Figure 17.** Accuracy by strategy and document length (left, Wilson bars) and by question (right).

**Table 15.** Accuracy of each strategy on the same 1120 documents per checkpoint (80 items, 1,000 and 1,800 tokens, seven evidence positions). Difference: paired accuracy change against head only with exact McNemar p, which treats the 14 documents of one item as independent (item-level intervals for the key contrasts are in the text).

| Checkpoint | Strategy | All | 1,000 tokens | 1,800 tokens | Difference vs head only | Cost |
|---|---|---|---|---|---|---|
| shipped | head only (default) | 28% [26-31] | 29% [25-33] | 28% [25-32] | n/a | 1 call, 512 |
| shipped | tail only | 32% [29-35] | 32% [28-36] | 32% [28-36] | +3.4 (p=0.021) | 1 call, 512 |
| shipped | head 25% + tail 75% | 33% [31-36] | 33% [29-37] | 33% [29-37] | +4.7 (p<0.001) | 1 call, 512 |
| shipped | head 50% + tail 50% | 34% [31-37] | 33% [30-37] | 35% [31-39] | +5.5 (p<0.001) | 1 call, 512 |

| Checkpoint | Strategy | All | 1,000 tokens | 1,800 tokens | Difference vs head only | Cost |
|---|---|---|---|---|---|---|
| shipped | lexical sentence selection | 32% [30-35] | 32% [28-36] | 33% [29-37] | +3.9 (p<0.001) | 1 call, 512 + text pass |
| shipped | window scan, max-confidence | 40% [38-43] | 43% [39-47] | 38% [34-42] | +11.9 (p<0.001) | 6.0 / 11.0 calls, 512 |
| shipped | max_len 1024 (no strategy) | 30% [28-33] | 31% [28-35] | 29% [25-33] | +1.7 (p=0.135) | 1 call, 1024 |
| shipped | max_len 2048 (no strategy) | 33% [30-35] | 32% [29-36] | 33% [29-37] | +4.1 (p<0.001) | 1 call, 2048 |
| v1 | head only (default) | 33% [31-36] | 34% [30-38] | 32% [29-36] | n/a | 1 call, 512 |
| v1 | tail only | 33% [30-36] | 32% [28-36] | 34% [30-38] | -0.3 (p=0.881) | 1 call, 512 |
| v1 | head 25% + tail 75% | 39% [36-42] | 39% [35-43] | 40% [36-44] | +5.9 (p<0.001) | 1 call, 512 |
| v1 | head 50% + tail 50% | 39% [36-42] | 39% [35-43] | 38% [34-42] | +5.4 (p<0.001) | 1 call, 512 |
| v1 | lexical sentence selection | 41% [38-44] | 42% [38-46] | 41% [37-45] | +7.9 (p<0.001) | 1 call, 512 + text pass |
| v1 | window scan, max-confidence | 46% [43-49] | 50% [46-54] | 43% [39-47] | +12.9 (p<0.001) | 6.0 / 11.0 calls, 512 |
| v1 | max_len 1024 (no strategy) | 36% [33-39] | 37% [33-41] | 35% [32-39] | +2.9 (p<0.001) | 1 call, 1024 |
| v1 | max_len 2048 (no strategy) | 39% [36-42] | 39% [35-43] | 40% [36-44] | +6.0 (p<0.001) | 1 call, 2048 |

For the checkpoints that cannot read a long document, almost every strategy beats the default and the window scan scores highest. For the released checkpoint the default scores 28%. The alternatives gain between 3 and 12 points, and the scan is best at 40%, +7.8 points above simply setting `max_len` to 2048 (p<0.001, n=1120; interval over items [+3.3, +12.9]). The short-row fine-tune behaves alike (scan 46%, 2048 window 39%, interval [+2.4, +12.0]), with lexical selection second (41%). The lead of the scan over the best other arrangement is smaller and less certain: +6.3 points [+0.9, +11.4] over head 50% + tail 50% for the released checkpoint and +5.1 [-1.1, +11.6] over lexical sentence selection for the short-row one. The scan size (312 tokens, stride 156) is the library default and was not tuned on these documents, but eight arrangements were compared on 80 items and no correction for multiplicity was applied, so the order among the weaker ones should not be over-read.

Our reading is that the gain comes from shorter windows and not from finding the evidence. Even the best strategy stays near the accuracy that these checkpoints reach on fully visible evidence in long documents (Section 4). The curves in Figure 16 show where each strategy wins: the default does well only when the evidence is at the very start, tail-only helps only at the very end, and head-plus-tail covers both ends but not the middle. The scan reads every part of the document in a 312-token window, where these models lose less to distraction than in a long one. Its accuracy is flat in the evidence position. It costs about 6 to 11 forward passes of 512 tokens (Section 9). Its max-confidence rule leans on a confidence that carries no information about whether evidence was seen (Section 6), so it is a heuristic that works on these checkpoints and not a principled retrieval. The long-row checkpoint was not run on these documents, so whether the scan still beats a long window for a checkpoint that can use one is open. The offset results suggest the larger window would then be competitive, but that is not a measurement.

## 9 What a larger window costs

If raising `max_len` were expensive the trade-off would be genuine, so we time it. One call of the released checkpoint (GPU: NVIDIA GeForce RTX 4090, bf16; weights resident: 1616 MB) is timed for two kinds of input at each setting: a short text (what most traffic is) and a text long enough to fill the window.

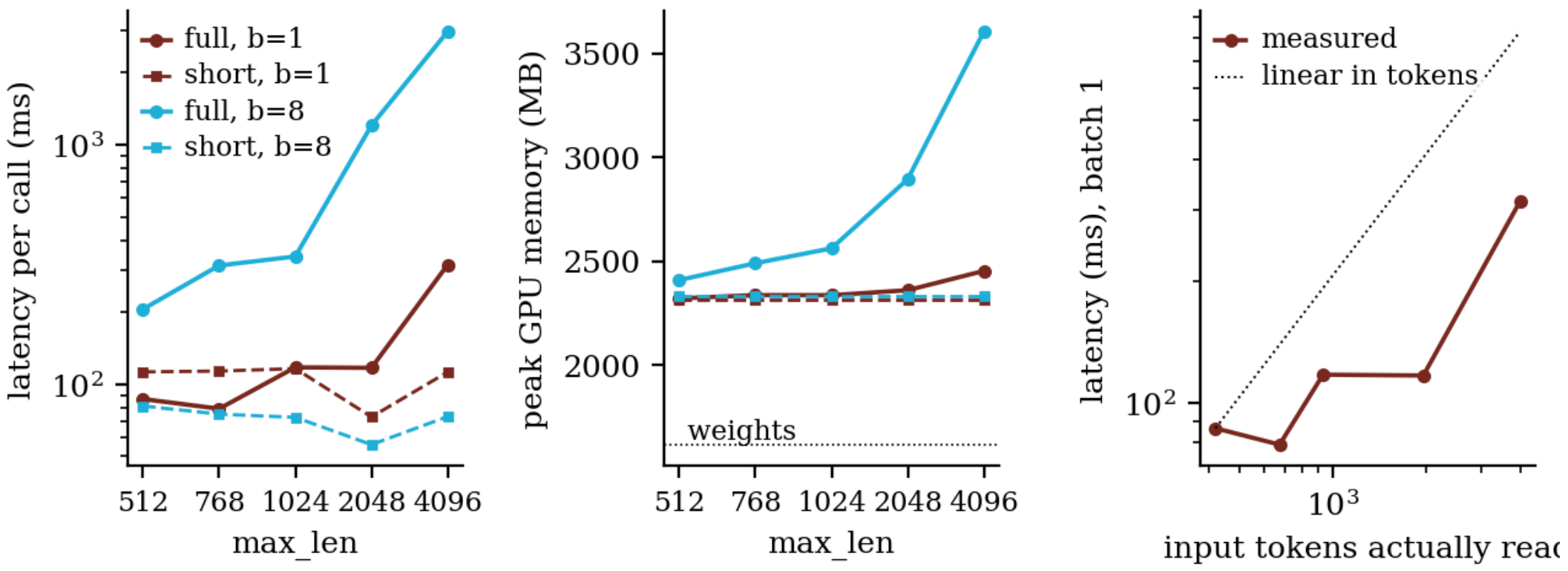

**Figure 18.** Latency and peak GPU memory of one call against `max_len` (released checkpoint; solid: the input fills the window, dashed: a short input), and latency against the number of tokens actually read.

**Table 16.** One call, median of five repetitions after one warm-up call on a shared GPU (latency is noisy; memory is not). 'Short': a sentiment text of a few dozen tokens; 'full': boilerplate that fills the window.

| max_len | short, b=1 (ms) | full, b=1 (ms) | tokens read | short, b=8 (ms) | full, b=8 (ms) | peak MB short b=1 | peak MB full b=1 | peak MB full b=8 |
|---|---|---|---|---|---|---|---|---|
| 512 | 113 | 87 | 419 | 81 | 206 | 2315 | 2320 | 2409 |
| 768 | 113 | 79 | 674 | 75 | 313 | 2315 | 2336 | 2490 |
| 1024 | 116 | 118 | 930 | 73 | 342 | 2315 | 2336 | 2561 |
| 2048 | 73 | 117 | 1955 | 56 | 1205 | 2315 | 2360 | 2894 |
| 4096 | 112 | 315 | 4003 | 73 | 2971 | 2315 | 2453 | 3605 |

For a short input the setting has no visible effect: batch-of-one latency is 113 ms at 512 and 112 ms at 4,096, and peak memory stays at 2315 and 2315 MB, because only the tokens present are processed. For an input that fills the window the cost grows with the tokens read, but a single request is dominated by fixed overhead and contention up to 2,048 tokens (87, 118 and 117 ms at 512, 1,024 and 2,048; 315 ms at 4,096). A batch of eight is more informative: 206 ms at 512, 1205 ms at 2,048 and 2971 ms at 4,096, which is 14.4 times the time for 9.6 times the tokens, a slightly super-linear growth. Peak memory for that batch is 2409 MB at 512 and 3605 MB at 4,096, against 1616 MB for the weights. A scan of a 1,800-token document in 512-token windows took 226 ms (11 windows), against 117 ms for one call at 2,048 on a similar number of tokens, so the scan is the more expensive way to read a long document.

The GPU was shared with other jobs while these numbers were taken, so absolute times are upper bounds and small differences between neighbouring settings are not meaningful. The memory figures are not affected by contention. For a service, a larger `max_len` costs nothing for short traffic and, for window-filling traffic, about what the extra tokens cost.

## 10 Which `max_len`? A guide

The experiments give a procedure that needs no retraining.

1. Tokenise a sample of the real states with the encoder's tokenizer and look at the upper percentiles. The snippet below does this in four lines, and the numbers it produces for our own corpus are in Table 17.
2. Set the window above the percentile you care about. Add the head (as large as 192 tokens when options have long descriptions, 26-67 tokens for the five questions here; Table 1), one separator, and a margin. Prefer a clean step (768, 1024, 2048) to an odd value, because the checkpoints were configured and trained at 512 or 1024 only, and for a checkpoint trained at 1,024 test any larger value first (Section 4, third feature). For the released checkpoint a larger window alone recovers little (Section 7).
3. Do not rely on confidence or on the flag to find the problem later. Count, on live traffic, the share of requests whose state exceeds the room. A service can alert on this number, and the runtime already records it.
4. For rows that still exceed the window, choose a strategy from Section 8 that matches where the decisive text usually sits (end of a thread, start of a report), or re-read with the scan.

5. If texts are long *and* the decisive span is a small part of them, test the checkpoint on documents of that shape. Section 7 shows that a checkpoint fine-tuned only on short rows can fail there even when nothing is cut.

```
from transformers import AutoTokenizer
import numpy as np
tok = AutoTokenizer.from_pretrained("answerdotai/ModernBERT-large")      # the encoder's tokenizer
lens = [len(tok(t, add_special_tokens=False)["input_ids"]) for t in my_states]
need = 192 + 1 + int(np.ceil(np.percentile(lens, 99)))                     # worst-case head: 192 tokens
max_len = next(L for L in (512, 768, 1024, 2048, 4096) if L >= need)
```

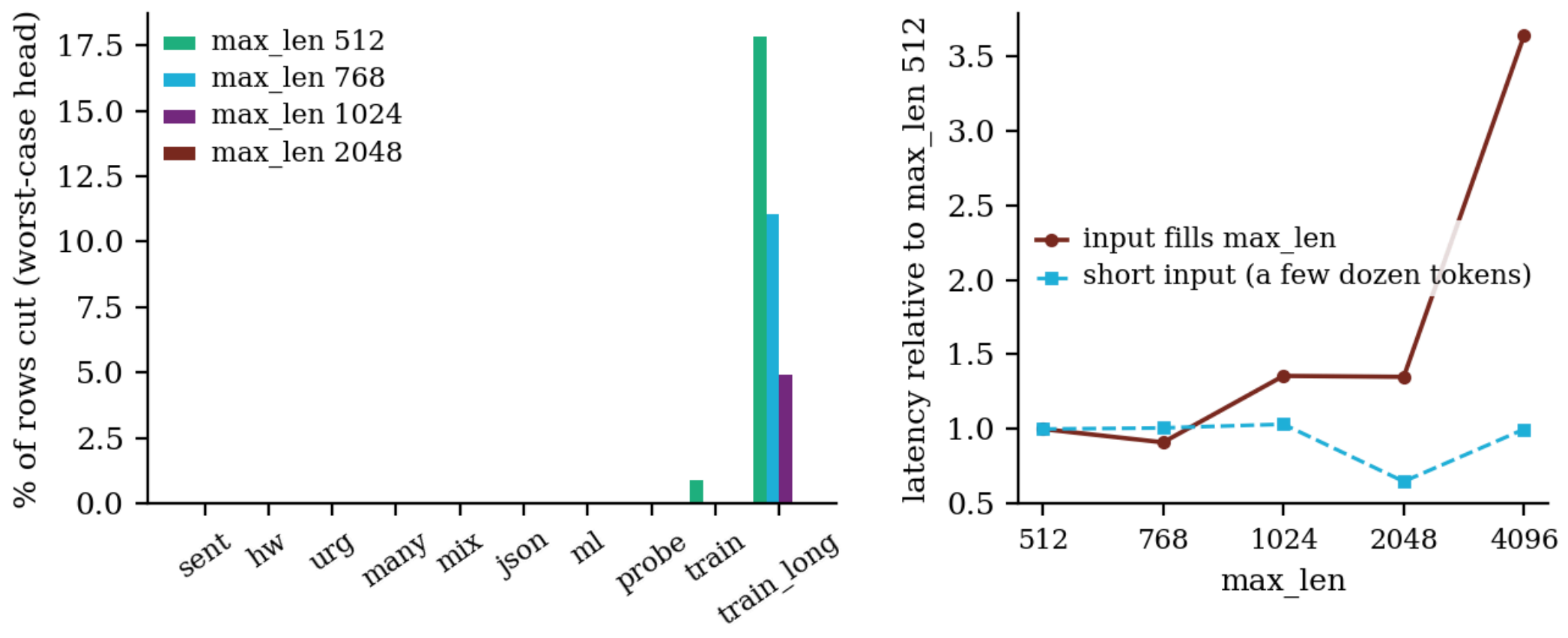

**Figure 19.** Left: share of rows cut at each `max_len` if the question head is at its 192-token cap (worst case). Right: cost of the larger window, as latency relative to 512, for short and window-filling inputs.

**Table 17.** `max_len` needed to avoid any cut for 95%, 99% and 100% of the rows of each data file, assuming the worst-case head of 192 tokens (needed = 192 + 1 + percentile of the state length), and the smallest of 512 / 768 / 1024 / 2048 / 4096 that provides it. The five shortest test files (sent, hw, urg, many, mix) need at most 275 tokens at every percentile, so 512 suffices for them.

| File | Rows | need 95% | need 99% | need 100% | setting 95% | setting 99% | setting 100% |
|---|---|---|---|---|---|---|---|
| json_test | 58 | 352 | 366 | 373 | 512 | 512 | 512 |
| ml_test | 322 | 361 | 388 | 404 | 512 | 512 | 512 |
| probe | 224 | 335 | 455 | 505 | 512 | 512 | 512 |
| train | 2,818 | 327 | 496 | 1324 | 512 | 512 | 2048 |
| train_long | 3,579 | 1024 | 1208 | 1382 | 1024 | 2048 | 2048 |

Costs are small where the corpus is mostly short. A larger `max_len` costs nothing for the short rows and costs latency only for the few rows that use the extra room (Section 9). With the corpus used here, 0.0% of the test rows are cut at 512 under the lightest head and 0.0% at 1,024. The default is therefore adequate for the bulk of the corpus, and the exposure is concentrated in a small set of long rows. Whether that holds for real traffic is the first thing to check.

## 11 Threats to validity, limitations and scorecard

The long documents are concatenations of a labelled row and unrelated sentences from small hand-written pools (62 and 30 sentences, reshuffled per item), spliced at a word boundary. Real long documents have structure (headings, quotations, repeated signatures, topic drift) that these lack, and a model may handle that structure better or worse. A control that splices at sentence boundaries is reported in Appendix D. The evidence rows are themselves short (median 25 tokens), so a decisive passage of several hundred tokens would behave differently at the cut; the straddling experiment is the best proxy.

Gold labels come from one LLM with a blind second pass and were cross-checked by another. Agreement of 94-99% means that a few percent of ‘errors’ at the ceiling may be label noise. This does not affect the comparisons between positions and window sizes, which use the same items throughout. We evaluate 177 items per experiment drawn evenly from the held-out sets, because the GPU was shared. Each task contributes at most 36 items. Per-cell intervals are wide (Wilson bands in the figures), and per-task differences of ten points should not be read as findings.

There is one training run per checkpoint and one stream per item, so the difference between the short-row and long-input checkpoints may include seed variation of a few points that we cannot separate. Latency was measured on a shared GPU and is an upper bound. The study covers English only and five questions. It tests one released model and two fine-tunes, with windows up to 4,096 tokens (2,048 in most experiments). It has no comparison with long-document architectures or with generative models. The relative-position and strategy experiments were run for the released and the short-row checkpoints only, because the GPU had no free memory for the remaining runs. The claims about the long-row checkpoint therefore rest on the offset, straddling, length, evidence-free, competing-evidence and distractor-type experiments. In particular we cannot say whether a window scan still beats a long window for it (Table F-1 lists what exists).

**Table 18.** Expectations E1 to E7 (Section 2.1) against the results. Numbers are released / short-row / long-row where three are given.

| ID | Expectation | Verdict | Evidence | Section |
|---|---|---|---|---|
| E1 | step in position | Partly | Sharp cliff in position (long-row: 75% at o=400, 28% at o=500 at 512), but the in-window level is not constant: 60% / 82% / 84% with evidence first against 38% / 41% / 68% pooled when visible | 4, 5 |
| E2 | step follows max_len and head size | Supported | Window ends sit at the room for 512, 768 and 1,024 (Table 4, Figure 5); per-question curves follow the question head (Figure 7) | 4 |
| E3 | confidence does not drop | Supported | AUROC for cut versus visible 0.50 / 0.53 / 0.55; mean confidence cut 0.57 / 0.67 / 0.82 | 6 |
| E4 | distractors, first piece wins | Partly | Boilerplate costs 27 points and other-task rows more, as expected; but with both pieces visible the *later* piece wins (released: 68% of adjacent sentiment pairs) | 5 |
| E5 | cost follows tokens read | Supported | Short input flat (113 and 112 ms); window-filling batch of 8: 206 ms at 512, 2971 ms at 4,096 | 9 |
| E6 | long training helps only if the baseline cannot use the window | Supported, except for look-alike text | Long-row minus short-row +27.6 points on visible evidence, +0.1 on cut; window change alone +6.2 for the released checkpoint; by distractor kind +19 (boilerplate), +13 (encyclopedic), -2 (other tasks' rows) | 7 |
| E7 | rearranging recovers some loss, no better than a larger window | Partly | Recovers up to 12 points; the window scan is +7.8 points above a 2,048 window for the released checkpoint, contrary to the expectation; long-row checkpoint not measured | 8 |

# 12 References and related work

We read the abstracts of eleven papers on the Hugging Face paper index (fetched on the day of writing), and every claim below about them is limited to what those abstracts state. The search was not exhaustive. Title-level queries for typed-decision, long-input and truncation terms turned up no paper on the input window of a typed-decision encoder, but a paper the search missed cannot be ruled out.

## 12.1 Position effects and long inputs in language models

Liu et al. [1] study decoder language models on multi-document question answering and key-value retrieval. They report that accuracy is highest when the relevant information is at the beginning or end of the input, drops when it is in the middle and falls as the input gets longer, even for models built for long contexts. Our setting differs in two ways that matter for interpretation. The model is a bidirectional encoder that reads the whole sequence at once, and the loss at the end is structural, because tokens beyond the window are absent from the computation. Where we measure how a visible position matters (Sections 4 and 5), the quantity is related but different, and we make no claim about the 'lost in the middle' U-shape for this model beyond what the figures show.

ModernBERT [2], the encoder inside Laya, is described as pre-trained with a native sequence length of 8,192 tokens and as efficient in speed and memory for inference on common GPUs. We use this only to motivate treating `max_len` as a configuration choice. The actual cost is measured on our hardware (Section 9), and we do not test the encoder beyond 4,096 tokens.

Levy et al. [7] add padding of different lengths, types and locations to the same reasoning questions and find that decoder accuracy degrades at input lengths far below the technical maximum. Their padding design is close to our distractor experiments (Section 5), but ours uses a bidirectional encoder with a fixed head and classification labels. Tworkowski et al. [11] name a 'distraction issue' for decoder models with external memory: as irrelevant keys outnumber relevant ones the model attends to the wrong ones, and they address this with contrastive training. Our long-row fine-tune is a much cruder cousin of that idea, and our distractor results (Section 5) measure a similar effect in an encoder without testing their method.

## 12.2 Long-document classification

Dai et al. [3] compare sparse-attention transformers with hierarchical ones (which split a long document into segments, encode them separately and combine them) on four document classification datasets. They observe a clear benefit from processing longer text and give advice on window size and splitting. Pappagari et al. [5] segment a long input into chunks, feed each to a BERT encoder and combine the outputs with a recurrent or transformer layer. Adhikari et al. [6] note that documents are often longer than the BERT input. Longformer [4] replaces full attention with a windowed pattern plus global tokens, so cost grows linearly with length. Our window-scan strategy (Section 8) is a training-free relative of the chunking idea, applied to a decision model whose options must be read together with every segment. The other strategies are inexpensive heuristics for a fixed-window encoder. We retrain no architecture and compare against none of these long-document models.

## 12.3 Typed-decision models

AnyJev [8] reads a typed decision from one prefill of a pretrained instruction-tuned language model and corrects a label prior and a bias towards positions in the option list without gradient steps. TypedBench [9] is a benchmark of calibration, framing sensitivity and cost for models that output probabilities over typed answers. Neither abstract mentions the length or truncation of the text being judged. Stekel [10] studies compact encoders for closed-set decisions and compares reading all options jointly (as Laya does) with scoring each option separately. The abstract reports that reordering the options changes 12 to 23% of the answers of the joint models and says nothing about input length. Our study is complementary: it concerns the *state* side of the input and not the option side, and it is specific to a fixed-window encoder.

## 12.4 What is and is not new

Three things are not new: that a fixed window loses what is outside it, that position matters in long inputs and that long-document classification needs longer, sparse or chunked encoders. To our knowledge, and within the limits of the search above, five things are new. They are a position-resolved measurement for one released typed-decision encoder, a test of whether confidence or the truncation flag warns of the loss, a paired comparison of long-input fine-tuning against a configuration change, training-free alternatives on identical documents and a latency and memory profile.

## Appendix A  Reproducing the study

Every experiment writes one JSON file per checkpoint into results (a8_<experiment>__<checkpoint>.json) and caches every distinct forward pass in an append-only line file, so an interrupted run resumes where it stopped. The GPU was shared and several runs were restarted automatically after driver faults, which is why the retry wrapper exists. The builder reads only these files and the earlier frozen probe results, and it recomputes every number and figure. Commands:

```
.\a8_run.ps1 base  "lengths,offset,null,partial,tsweep,duel,dilute,frac,strategy,cost,libscan"   # released checkpoint
.\a8_run.ps1 v1    "offset,null,partial,tsweep,duel,dilute,frac,strategy"
.\a8_run.ps1 v1_long "offset,null,partial,tsweep,duel,dilute,frac,strategy"
python a8_exp.py splice base                                                    # splice control (Appendix D)
.\.venv\Scripts\python.exe build_long.py a8                                      # figures, tables and this document
```

The SHA-256 of the released-checkpoint offset result begins 5317404dd0e7a081. Distractor sentences are in a8_pools.py (62 boilerplate and 30 encyclopedic sentences); each item's stream is a seeded shuffle of the pool, so the documents are deterministic.

## Appendix B  Examples of constructed documents

The same evidence (one sentiment row, gold label sentiment index 0) at three offsets of an 1,800-token document. The text shown is 90 characters on each side of the evidence (bold); the rest is boilerplate from the same stream.

**Table B-1.** Three constructed documents.

| o | Tokens | Evidence tokens | Context |
|---|---|---|---|
| 0 | 1800 | 0-10 | **Wow. Just wow. Incredible service.** Receipts for taxi rides need to show the date, the route and the amount paid. Please retu … |
| 200 | 1800 | 200-210 | al opening hours for the cafeteria are announced on the digital signs near the lifts. All **Wow. Just wow. Incredible service.** printers on the second floor support double sided printing, scanning to email and secure … |
| 1200 | 1800 | 1200-1210 | dings, shuttle stops and bicycle racks is available at reception. The archive holds paper **Wow. Just wow. Incredible service.** records going back to the nineteen seventies, indexed by year and department. The recepti … |

## Appendix C  Accuracy by offset and window, all checkpoints

**Table C-1.** Accuracy (%) by evidence offset o (columns) for every checkpoint and max_len; n = 177 requests per cell.

| Checkpoint | max_len | 0 | 100 | 200 | 300 | 400 | 500 | 600 | 800 | 1000 | 1500 |
|---|---|---|---|---|---|---|---|---|---|---|---|
| shipped | 512 | 59 | 33 | 36 | 34 | 36 | 27 | 27 | 27 | 27 | 27 |
| shipped | 768 | 62 | 32 | 36 | 35 | 36 | 32 | 36 | 26 | 26 | 26 |
| shipped | 1024 | 59 | 34 | 36 | 36 | 34 | 32 | 33 | 32 | 27 | 27 |
| shipped | 2048 | 60 | 38 | 39 | 40 | 40 | 36 | 36 | 36 | 33 | 33 |
| v1 | 512 | 69 | 34 | 36 | 34 | 55 | 27 | 27 | 27 | 27 | 27 |
| v1 | 768 | 73 | 36 | 38 | 35 | 36 | 32 | 33 | 28 | 28 | 28 |
| v1 | 1024 | 73 | 35 | 38 | 34 | 36 | 32 | 34 | 33 | 27 | 27 |
| v1 | 2048 | 82 | 34 | 38 | 40 | 38 | 34 | 37 | 36 | 30 | 29 |
| v1_long | 512 | 77 | 67 | 67 | 64 | 75 | 28 | 28 | 28 | 28 | 28 |
| v1_long | 768 | 79 | 67 | 64 | 62 | 67 | 62 | 67 | 27 | 27 | 27 |
| v1_long | 1024 | 80 | 68 | 64 | 62 | 69 | 62 | 63 | 66 | 27 | 27 |
| v1_long | 2048 | 84 | 74 | 75 | 74 | 76 | 69 | 75 | 71 | 53 | 53 |

## Appendix D  Splice check: word boundary versus sentence boundary

The main experiments cut the stream at a word boundary, so the evidence sometimes lands in the middle of a boilerplate sentence. To check that this is not what drives the results we repeated a 900-token document with the evidence spliced at the next sentence boundary (released checkpoint, `max_len` 2048, all 177 items).

**Table D-1.** Accuracy by splice rule.

| o | Word boundary | Sentence boundary | McNemar p |
|---|---|---|---|
| 100 | 32.8% [26.3-40.0] | 35.0% [28.4-42.3] | p=0.219 |
| 300 | 33.9% [27.3-41.1] | 35.6% [28.9-42.9] | p=0.508 |

## Appendix E  Window scan: our re-implementation versus the library call

The window scan in Section 8 re-implements the library's long-input method so that windows can be batched and shared. As a check we ran both on documents spread over the item list (1,800 tokens, evidence at offset 800, `max_len` 512).

**Table E-1.** Agreement of the two implementations.

| Documents | Same top answer | Windows (library) | Windows (ours) |
|---|---|---|---|
| 13 | 100.0% | 11.0 | 11.0 |

## Appendix F  Experiment sizes

**Table F-1.** Number of scored requests per experiment and checkpoint (before sharing of identical visible prefixes).

| Experiment | shipped | v1 | v1_long |
|---|---|---|---|
| offset | 7,080 | 7,080 | 7,080 |
| null | 3,960 | 3,960 | 3,960 |
| partial | 2,124 | 2,124 | 2,124 |
| tsweep | 4,248 | 4,248 | 4,248 |
| duel | 1,920 | 1,920 | 1,920 |
| dilute | 4,173 | 4,173 | 4,173 |
| frac | 3,540 | 3,540 | n/a |
| strategy | 7,840 | 7,840 | n/a |